%% file: iclr2021_conference.tex
\documentclass{article} 
\usepackage{iclr2021_conference,times}

\usepackage{epsfig}
\usepackage{graphicx}
\usepackage{amsmath}
\usepackage{amssymb}
\usepackage{setspace}
\usepackage{wrapfig}
\usepackage{floatrow}
\usepackage{array}
\usepackage{pythonhighlight}
\usepackage{natbib}
\usepackage{wrapfig}
\usepackage{etoolbox}
\usepackage{pgf}
\usepackage{array}

\usepackage{inconsolata}

\usepackage{algorithm}
\usepackage{algpseudocode}

\usepackage{multirow}
\usepackage{rotating}
\usepackage{booktabs}

\usepackage{enumitem}
\usepackage[olditem,oldenum]{paralist}

\usepackage{alltt}
\usepackage{listings}

\usepackage{standalone/mysymbols}

\usepackage{url}
\usepackage{xspace}
\usepackage{comment}
\usepackage{color}
\usepackage{xcolor}
\usepackage{afterpage}
\usepackage{pdfpages}
\usepackage{framed}
\usepackage{fancybox}
\usepackage{cuted}
\usepackage{caption}
\usepackage{subcaption}
\usepackage{paralist}

\usepackage{footnote}
\makesavenoteenv{tabular}

\definecolor{citecolor}{HTML}{2779af}
\definecolor{linkcolor}{HTML}{c0392b}
\usepackage[pagebackref=false,breaklinks=true,colorlinks,bookmarks=false,citecolor=citecolor,linkcolor=linkcolor]{hyperref}

\newcommand\tstrut{\rule{0pt}{2.4ex}}
\newcommand\bstrut{\rule[-1.0ex]{0pt}{0pt}}

\input{standalone/definitions}
\input{standalone/space_saver}

\title{Large Batch Simulation\\for Deep Reinforcement Learning}

\author{
    \begin{tabular}[h]{l}
        Brennan Shacklett$^1$\thanks{Correspondence to bps@cs.stanford.edu}~~
        Erik Wijmans$^2$~
        Aleksei Petrenko$^{3,4}$ ~ \\
        Manolis Savva$^5$~
        Dhruv Batra$^2$~
        Vladlen Koltun$^3$~
        Kayvon Fatahalian$^1$
    \end{tabular}\\
    \begin{tabular}[h]{l}
        $^1$Stanford University ~
        $^2$Georgia Institute of Technology ~
        $^3$Intel Labs ~ \\
        $^4$University of Southern California ~ 
        $^5$Simon Fraser University ~
    \end{tabular}\\
}

\iclrfinalcopy 
\begin{document}

\maketitle

\begin{abstract}
    We accelerate deep reinforcement learning-based training in visually complex 3D environments by two orders of magnitude over prior work, realizing end-to-end training speeds of over 19,000 frames of experience per second on a \emph{single} GPU and up to 72,000 frames per second on a single eight-GPU machine.  The key idea of our approach is to design a 3D renderer and embodied navigation simulator around the principle of ``batch simulation'': accepting and executing large batches of requests simultaneously.
    Beyond exposing large amounts of work at once, batch simulation allows implementations to amortize in-memory storage of scene assets, rendering work, data loading, and synchronization costs across many simulation requests, dramatically improving the number of simulated agents per GPU and overall simulation throughput.  
    To balance DNN inference and training costs with faster simulation, we also build a computationally efficient policy DNN that maintains high task performance, and modify training algorithms to maintain sample efficiency when training with large mini-batches.
    By combining batch simulation and DNN performance optimizations, we demonstrate that PointGoal navigation agents can be trained in complex 3D environments on a single GPU in 1.5 days to 97\% of the accuracy of agents trained on a prior state-of-the-art system using a 64-GPU cluster over three days. We provide open-source reference implementations of our batch 3D renderer and simulator to facilitate incorporation of these ideas into RL systems.

\end{abstract}

\vspace{-0.1in}
\csection{Introduction}

Speed matters.
It is now common for modern reinforcement learning (RL) algorithms leveraging deep neural networks (DNNs) to require
\emph{billions} of samples of experience from simulated environments 
\citep{ddppo,petrenko2020sample,openai2019dota,silver2017mastering,alphastarblog}.
For embodied AI tasks such as visual navigation, 
where the ultimate goal for learned policies is deployment in the real world,
learning from \emph{realistic simulations} is important for successful transfer of learned policies to physical robots.
In these cases simulators must render detailed 3D scenes and simulate agent interaction  with complex environments~\citep{ai2thor,dosovitskiy2017carla,habitat19iccv,xia2020interactive,tdw}.

Evaluating and training a DNN on billions of simulated samples
is computationally expensive.
For instance, 
the DD-PPO system~\citep{ddppo} used 64~GPUs over three days to learn 
from 2.5~billion frames of experience and achieve near-perfect PointGoal navigation 
in 3D scanned environments of indoor spaces.
At an even larger distributed training scale, OpenAI Five used over 50,000~CPUs and 1000~GPUs to train Dota 2 agents~\citep{openai2019dota}.
Unfortunately, experiments at this scale are out of reach for most researchers. This problem will only grow worse
as the field explores more complex tasks in more detailed environments.

Many efforts to accelerate deep RL focus on improving the efficiency of DNN evaluation and training
-- \eg, by ``centralizing'' computations to facilitate efficient
batch execution on GPUs or TPUs~\citep{espeholt2020seed,petrenko2020sample} or by parallelizing across GPUs~\citep{ddppo}.
However, most RL platforms still accelerate environment simulation by running \emph{many copies of off-the-shelf, unmodified simulators}, 
such as simulators designed for video game engines~\citep{bellemare2013arcade,kempka2016vizdoom,beattie2016deepmind,AllenAct},
on large numbers of CPUs or GPUs. This approach is a simple and productive way to improve simulation throughput, 
but it makes inefficient use of computation resources.
For example, when rendering complex environments~\citep{ai2thor,habitat19iccv,xia2018gibson},
a single simulator instance might consume gigabytes of GPU memory,
limiting the total number of instances to far below the parallelism afforded by the machine.
Further, running many simulator instances (in particular when they are distributed across machines) can introduce overhead in synchronization and communication with other components of the RL system.
Inefficient environment simulation is a major reason RL platforms typically require scale-out parallelism to achieve high end-to-end system throughput.

In this paper, we crack open the simulation black box and take a holistic approach to co-designing a 3D renderer, simulator, and RL training system.
Our key contribution is \emph{batch simulation} for RL: designing high-throughput simulators that accept large batches of 
requests as input (aggregated across different environments, potentially with different assets) and 
efficiently execute the entire batch at once. 
Exposing work en masse facilitates a number of optimizations: 
we reduce memory footprint by sharing scene assets (geometry and textures) across rendering requests 
(enabling orders of magnitude more environments to be rendered simultaneously on a single GPU), 
amortize rendering work using GPU commands that draw triangles from multiple scenes at once, hide latency of scene I/O, 
and exploit batch transfer to reduce data communication and synchronization costs between the simulator, DNN inference, and training. 
To further improve end-to-end RL speedups, the DNN workload
must be optimized to match high simulation throughput,
so we design a computationally efficient policy DNN that still achieves high task performance in our experiments.
Large-batch simulation increases the number of samples collected per training iteration, 
so we also employ techniques from large-batch supervised learning to maintain sample efficiency in this regime.  

We evaluate batch simulation on the task of PointGoal navigation~\citep{anderson2018evaluation} in 3D scanned 
Gibson and \matterport environments, and show that end-to-end optimization of batched rendering, simulation, inference, 
and training yields a 110$\times$ speedup over state-of-the-art prior systems, 
while achieving 97\% of the task performance for depth-sensor-driven agents and 91\% for RGB-camera-driven agents. 
Concretely, we demonstrate sample generation and training at over 19,000 frames of experience per second on a single GPU.%
\footnote{Samples of experience used for learning, not `frameskipped' metrics typically used in Atari/DMLab.}
In real-world terms, a single GPU is capable of training a virtual agent on 26 years of experience in a single day.%
\footnote{Calculated on rate a physical robot 
(LoCoBot~\citep{locobot}) 
 collects observations when 
operating constantly at maximum speed ($0.5$ m/s) and capturing $1$ frame every $0.25$m.}
This new performance regime significantly improves the accessibility and efficiency of RL research in realistic 3D environments, and opens new possibilities for more complex embodied tasks in the future.

\begin{figure}
\vspace{-3em}
    \includegraphics[width=\linewidth]{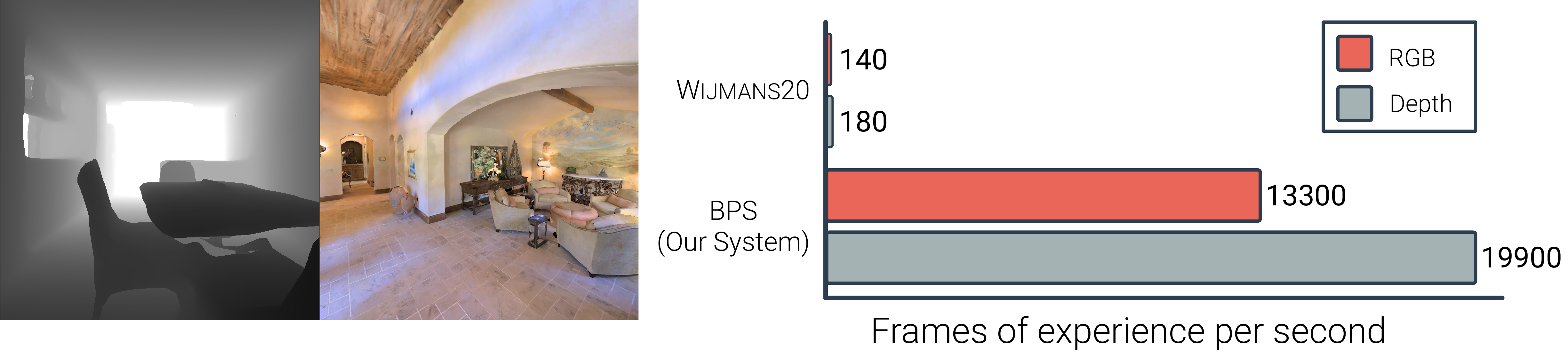}
    \vspace{1mm}
    \caption{We train agents to perform \pointnavfull in visually complex Gibson~\citep{xia2018gibson} and \matterport~\citep{mp3d} environments such as the ones shown here. These environments feature detailed scans of real-world scenes composed of up to 600K triangles and high-resolution textures. Our system is able to train agents using 64$\times$64 depth sensors (a high-resolution example is shown on the left) in these environments at 19,900 frames per second, and agents with 64$\times$64 RGB cameras at 13,300 frames per second on a \emph{single} GPU.}
    \label{fig:teaser}
    \vspace{-2em}
\end{figure}

\input{sections/main/related-work}

\csection{System Design \& Implementation}\label{sec:impl}

\begin{figure}[t]
\vspace{-3em}
\centering
\includegraphics[width=\linewidth]{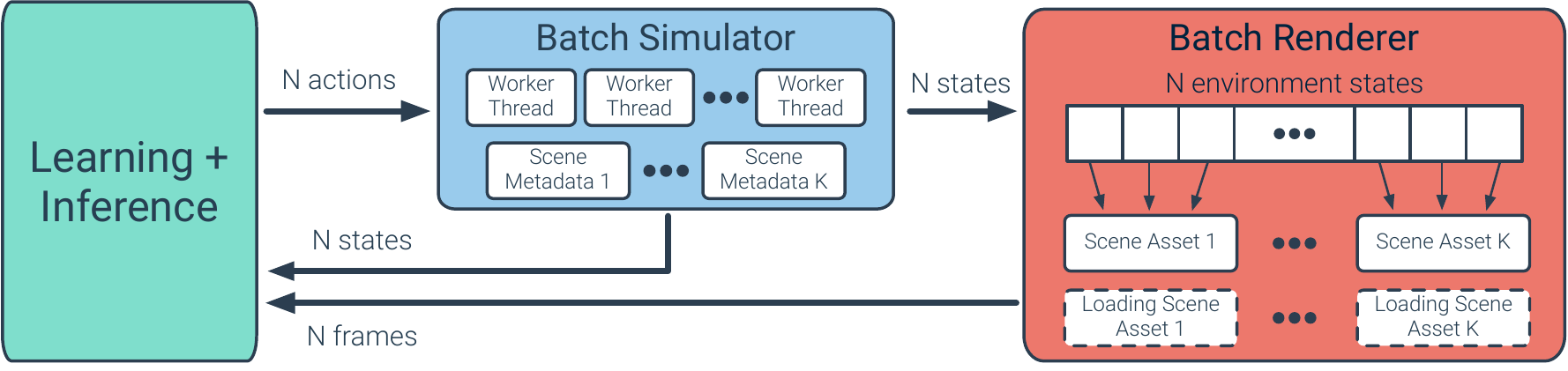}
\vspace{-2mm}
\caption{The batch simulation and rendering architecture. Each component communicates at the granularity of batches of $N$ elements (e.g., $N$=1024), minimizing communication overheads and allowing components to independently parallelize their execution over each batch. 
To fit the working set for large batches on the GPU, the renderer maintains 
$K\! \ll\! N$ unique scene assets in GPU memory and shares these assets across subsets of the $N$ environments in a batch. 
To enable experience collection across a diverse set of environments, the renderer continuously updates the set of $K$ in-memory scene assets using asynchronous transfers that overlap rollout generation and learning.
}
\label{fig:architecture}
\vspace{-5mm}
\end{figure}

Batch simulation accelerates rollout generation during RL training by processing many simulated environments simultaneously in large batches. \figref{fig:architecture} illustrates how batch simulation interacts with policy inference to generate rollouts. Simulation for sensorymotor agents, such as the \pointnavfull task targeted by our implementation, can be separated into two tasks: determining the next environment state given an agent's actions and rendering its sensory observations. Therefore, our design utilizes two components: a batch simulator that performs geodesic distance and navigation mesh \citep{navmesh} computations on the CPU, and a batch renderer that renders complex 3D environments
on the GPU.

During rollout generation, batches of requests are passed between these components -- given $N$ agents, the simulator produces a batch of $N$ environment states. 
Next, the renderer processes the batch of environment states by simultaneously rendering $N$ frames and exposing the result directly in GPU memory.  Agent observations (from both the simulator and the renderer) are then provided as a batch to policy inference
to determine the next actions for the $N$ agents.

The key idea is that the batch simulator and renderer implementations
(in addition to the DNN workload)
\emph{take responsibility for their own parallelization}.   
Large batch sizes (values of $N$ on the order of hundreds to thousands of environments) 
provide opportunities for implementations to efficiently utilize parallel execution resources (e.g., GPUs) as well as amortize processing, synchronization, and data communication costs across many environments.
The remainder of this section describes the design and key implementation details of our system's batch simulator and batch renderer,
as well as contributions that improve the efficiency of policy inference and optimization in this regime.

\csubsection{Batch Environment Simulation}

Our CPU-based batch simulator executes geodesic distance and navigation mesh computations in parallel for a large batch of environments. Due to differences in navigation mesh complexity across environments, the time to perform simulation may differ per environment. This variance is the source of workload imbalance problems in parallel synchronous RL systems~\citep{ddppo,habitat19iccv} and one motivation for recent asynchronous designs~\citep{petrenko2020sample,espeholt2020seed,espeholt2018impala}.
To ensure good workload balance, our batch simulator operates on large batches that contain \emph{significantly more} environments than the number of available CPU cores and dynamically schedules work onto cores using a pool of worker threads (simulation for each environment is carried out sequentially).
Worker threads report simulation results into a designated per-environment slot in a results buffer that is 
communicated to the renderer via a single batched request when all environment simulation for a batch is complete. To minimize CPU memory usage, the simulator only loads navigation meshes and does not utilize the main rendering assets.

\csubsection{Batch Rendering}

A renderer for producing RL agent observations in scanned real-world environments must efficiently synthesize many low-resolution renderings (e.g., 64$\times$64 pixels) of scenes featuring high-resolution textures and complex meshes.\footnote{The Matterport3D dataset contains up to 600K triangles per 3D scan.}
Low-resolution output presents challenges for GPU acceleration.  
Rendering images one at a time produces too little rendering work to efficiently utilize a modern GPU rendering pipeline's parallel processing resources. Rendering many environments concurrently but individually (e.g., from different worker threads or processes) exposes more rendering work to the GPU,
but incurs the overhead of sending the GPU many fine-grained rendering commands.

To address the problem of rendering many small images efficiently, our renderer combines the GPU commands required to render observations for an entire simulation batch of $N$ environments into a \emph{single rendering request} to the GPU -- effectively drawing the entire batch as a single large frame (individual environment observations are tiles in the image).
This approach exposes large amounts of rendering work to the GPU and amortizes GPU pipeline configuration and rendering overhead over an entire batch.  Our implementation makes use of modern GPU pipeline features~\citep{khronos2017vulkan} that allow rendering tasks that access different texture and mesh assets to proceed as part of a single large operation (avoiding GPU pipeline flushes due to pipeline state reconfiguration).

\textbf{Scene asset sharing.} Efficiently utilizing a GPU requires batches to be large (we use $N$ up to 1024).
However, geometry and texture assets for a single environment may be gigabytes in size,
so naively loading unique assets for each environment in a large batch would exceed available GPU memory.  Our implementation allows multiple environments in a batch to reference the same 3D scene assets in GPU memory. Specifically, our system materializes $K$ unique assets in GPU memory ($K\! \ll\! N$) and constructs batches of $N$ environments that reference these assets.
Asset reuse decreases the diversity of training experiences in a batch, so to preserve diversity we limit the ratio of $N$ to $K$ in any one batch to 32, and continuously rotate the set of $K$ assets in GPU memory. The renderer refreshes the set of $K$ assets by asynchronously loading new scene assets into GPU memory during the main rollout generation and learning loop.
As episodes complete, new environments are constructed to reference the newly loaded assets, and assets no longer referenced by active environments are removed from GPU memory. This design allows policy optimization to learn from an entire dataset of assets without exceeding GPU memory or incurring the latency costs of frequent asset loading.

\textbf{Pipelined geometry culling.} When rendering detailed geometry to low-resolution images,
most scene triangles cover less than one pixel.
As a result, rendering performance is determined by the rate the GPU's rasterization hardware processes triangles, not the rate the GPU can shade covered pixels.  To reduce the number of triangles the GPU pipeline must process,
the renderer uses idle GPU cores to identify and discard geometry that lies outside the agent's view---a process known as frustum culling~\citep{akenine2019real}.
Our implementation pipelines frustum culling operations (implemented using GPU compute shaders) with rendering for different environments in a batch. This pipelined design increases GPU utilization by concurrently executing culling work on the GPU's programmable cores and rendering work on the GPU's rasterization hardware. 

\csubsection{Policy DNN Architecture}

High-throughput batch simulation creates a need for high-throughput policy DNN inference.
Therefore, we develop a policy DNN architecture designed to achieve an efficient balance between high task performance and low computational cost. Prior work in \pointnavfull \citep{ddppo} used a policy DNN design where a visual encoder CNN processes an agent's visual sensory information followed by an LSTM~\citep{hochreiter97lstm} that determines the policy's actions. Our policy DNN uses this core design augmented with several performance optimizations.

First, we reduce DNN effective input resolution from 128$\times$128~\citep{ddppo} to 64$\times$64. 
Beyond this simple optimization, we choose a shallow visual encoder CNN -- a nine-layer ResNet~\citep{he2016resnet} (ResNet18 with every other block removed), rather than the 50 layer (or larger) ResNets used by prior work. To counteract reduced task performance from the ResNet's relatively low capacity, all stages include Squeeze-Excite (SE) blocks~\citep{hu2018senet} with $r{=}16$. Additionally, we use a \texttt{SpaceToDepth} stem~\citep{ridnik2020tresnet}, which we find performs equally to the standard \texttt{Conv+MaxPool} stem while using less GPU memory and compute. 

Finally, we avoid the use of normalization layers in the ResNet as these require spatial reductions over the feature maps, preventing layer-fusion optimizations.
Instead, the CNN utilizes Fixup Initialization~\citep{zhang2019fixup} to improve training stability.
Fixup Initialization replaces expensive normalization layers after each convolution with cheap elementwise multiplication and addition.

\csubsection{Large Mini-Batch Policy Optimization}

In on-policy reinforcement learning, policy optimization utilizes trajectories of experience to reduce  bias and for backpropagation-through-time.  When generating trajectories of length $L$ with a simulation batch size of $N$, a rollout will have $N{\times}L$ steps of experience.   
Therefore, a consequence of simulation with large $N$ is that more experience is collected per rollout.

Large $N$ presents the opportunity to utilize large mini-batches to improve the throughput of policy optimization; however, throughput must be balanced against generalization and sample efficiency to ensure that reduced task performance does not offset the throughput gains. Although large mini-batch training is known to hurt generalization in supervised learning~\citep{keskar2016large}, we do not see evidence of this for RL. Conversely, we do find that sample efficiency for \pointnavfull is harmed by naively increasing $N$. Fortunately, we are able to mitigate this loss of sample efficiency using techniques for improving generalization from the large mini-batch optimization literature.

First, we scale the learning rate by $\sqrt{\frac{B}{B_{\text{base}}}}$, where $B_{\text{base}}{=}256$ and $B$, the training batch size, is $N{\times}L$ divided by the number of mini-batches per training iteration. We find it beneficial to use the scaled learning rate immediately instead of `warming-up' the learning rate~\citep{goyal2017accurate}. Second, we use and adapt the Lamb optimizer~\citep{you2020large}.    Lamb is a modification to Adam~\citep{kingma2016adam} that applies LARS~\citep{you2017scaling} to the step direction estimated by Adam to better handle high learning rates. Since the Adam optimizer is often used with PPO~\citep{schulman2017ppo}, Lamb is a natural choice.
Given the Adam step direction $s^{(k)}_t$ for weights $\theta^{(k)}_t$,
\begin{align}
  \theta^{(k)}_{t+1} &= \theta^{(k)}_t - \eta_t r^{(k)}_t (s^{(k)}_t + \lambda \theta^{(k)}_t) &
  r^{(k)}_t &= \frac{\phi(||\theta^{(k)}_t||)}{||s^{(k)}_t + \lambda \theta^{(k)}_t||}
\end{align}
where $\eta_t$ is the learning rate and $\lambda$ is the weight decay coefficient.  We set $\phi(||\theta^{(k)}_t||)$ as $\min\{||\theta^{(k)}_t||, 10.0\}$  and introduce an additional clip on the trust ratio $r^{(k)}_t$:
\begin{align}
    r^{(k)}_t &= \min\left\{\max\left\{\frac{\phi(||\theta^{(k)}_t||)}{||s^{(k)}_t + \lambda \theta^{(k)}_t||}, \rho \right\},\frac{1}{\rho}\right\}
\end{align}
We find the exact value of $\rho$ to be flexible (we observed similar training with ${\rho\in \{10^{-2}, 10^{-3}, 10^{-4}\}}$) and also observed that this clip is only influential at the start of training, suggesting that there is an initialization scheme where it is unnecessary.

\csection{Results}
We evaluate the impact of our contributions on end-to-end training speed and task performance by training \pointnavfull agents in the complex Gibson~\citep{xia2018gibson} and \matterport~\citep{mp3d} environments.  The fastest published end-to-end training performance in these environments is achieved with the synchronous RL implementation presented with DD-PPO \citep{ddppo}. Therefore, both our implementation and the baselines we compare against are synchronous PPO-based RL systems.

\csubsection{Experimental Setup}\label{sec:setup}
\xhdr{\pointnavfull task.} We train and evaluate agents via the same procedure as~\citet{ddppo}: agents are trained for \pointnav~\citep{anderson2018evaluation} with either a \depth sensor or an \rgb camera. \depth agents are trained on Gibson-2plus~\citep{xia2018gibson} and, consistent with \citet{ddppo}, \rgb agents are also trained on \matterport~\citep{mp3d}. \rgb camera simulation requires textures for the renderer, increasing the GPU memory consumed by each scene significantly. Both classes of agent are trained on 2.5 billion simulated samples of experience. 

Agents are evaluated on the Gibson dataset~\citep{xia2018gibson}. We use two metrics: Success, whether or not the agent reached the goal, and SPL~\citep{anderson2018evaluation}, a measure of both Success and efficiency of the agent's path. We perform policy evaluation using  Habitat-Sim~\citep{habitat19iccv}, unmodified for direct comparability to prior work.

\xhdr{Batch Processing Simulator (\bps).}
We provide an RL system for learning \pointnav built around the batch simulation techniques and system-wide optimizations described in Section~\ref{sec:impl}. The remainder of the paper refers to this system as \bps (Batch Processing Simulator). To further accelerate the policy DNN workload,
\bps uses half-precision inference and mixed-precision training.

\xhdr{Baseline.}
The primary baseline for this work is \citet{ddppo}'s open-source \pointnav implementation, which uses Habitat-Sim~\citep{habitat19iccv} -- the prior state of the art in high-performance simulation of realistic environments such as Gibson. Unlike \bps, multiple environments are simulated simultaneously using parallel worker processes that render frames at 256$\times$256 pixels before downsampling to 128$\times$128 for the visual encoder. The fastest published configuration uses a ResNet50 visual encoder. Subsequent sections refer to this implementation as \baseline. 

\xhdr{Ablations.}
As an additional baseline, we provide \baselinepp, which uses the optimized SE-ResNet9-based policy DNN (including performance optimizations and resolution reduction relative to \baseline) developed for \bps, but otherwise uses the same system design and simulator as \baseline (with a minor modification to not load textures for \depth agents). \baselinepp serves to isolate the impact of two components of \bps: first, the low-level DNN efficiency improvements, and, more importantly, the performance of batch simulation versus \baseline's independent simulation worker design. Additionally, to ablate the effect of our encoder CNN architecture optimizations, we include a variant of \bps, \bpsBIG, that uses the same ResNet50 visual encoder and input resolution as \baseline, while maintaining the other of optimizations \bps.

\xhdr{Multi-GPU training.} 
To support multi-GPU training, all three systems replace standard PPO with DD-PPO~\citep{ddppo}. DD-PPO scales rollout generation and policy optimization across all available GPUs, scaling the number of environments simulated and the number of samples gathered between training iterations proportionally. We report results with eight GPUs.

\xhdr{Determining batch size.} The per-GPU batch size, $N$, controls a trade-off between memory usage, sample efficiency, and speed.
For \bps, $N$ designates the batch size for simulation, inference, and training.
For \baseline and \baselinepp, $N$ designates the batch size for inference and training,
as well as the number of simulation processes.
\baseline sets $N$=$4$ for consistency with \citet{ddppo}.
To maximize performance of single-GPU runs, \bps uses the largest batch size that fits in GPU memory,
subject to the constraint that no one scene asset can be shared by more than 32 environments in the batch.
In eight-GPU configurations, DD-PPO scales the number of parallel rollouts with the number of GPUs, so to maintain reasonable sample efficiency \bps limits per-GPU batch size to $N$=$128$, with $K$=$4$ active scenes per GPU.
\baselinepp \depth experiments use $N{=}64$ (limited by system memory due to $N$ separate processes running Habitat-Sim).
Batch size in \baselinepp \rgb experiments
is limited by GPU memory ($N$ ranges from 6 to 20 depending on the GPU).
Appendix~\ref{experiment_appendix} provides the batch sizes used in all experiments.

\xhdr{Benchmark evaluation.}
We report end-to-end performance benchmarks in terms of average frames per second (FPS) achieved by each system. We measure FPS as the number of samples of experience processed over 16,000 inference batches divided by the time to complete rollout generation and training for those samples. In experiments that run at 128$\times$128 pixel sensor resolution, rendering occurs at 256$\times$256 and is downsampled for the policy DNN to match the behavior of \baseline regardless of system, while 64$\times$64 resolution experiments render without downsampling. Results are reported across three models of NVIDIA GPUs: Tesla V100, GeForce RTX~2080Ti, and GeForce RTX~3090. (The different GPUs are also accompanied by different CPUs, see Appendix~\ref{hwappendix}.)

\begin{table}[t]
\vspace{-2em}
\centering
\setlength{\tabcolsep}{4pt}
\resizebox{\textwidth}{!}{
\begin{tabular}{c lr r r r r r r}
    \toprule
    \multicolumn{1}{c}{\multirow[b]{2}{*}{\bf Sensor }} & \multicolumn{1}{c}{\multirow[b]{2}{*}{\bf System }} &
    \multicolumn{1}{c}{\multirow[b]{2}{*}{\bf CNN }} &
    \multicolumn{1}{c}{\multirow[b]{2}{*}{\bf \shortstack[c]{Agent\\Res.} }} &
    \multicolumn{1}{c}{\multirow[b]{2}{*}{\bf RTX 3090}} &
    \multicolumn{1}{c}{\multirow[b]{2}{*}{\bf RTX 2080Ti}} & \multicolumn{1}{c}{\multirow[b]{2}{*}{\bf Tesla V100}} &  \multicolumn{1}{c}{\multirow[b]{2}{*}{\bf 8$\times$2080Ti}} & \multicolumn{1}{c}{\multirow[b]{2}{*}{\bf 8$\times$V100}}\tstrut\\
    &&&&&&&&\\
    \midrule 
    \multirow{4}{*}{\depth}
    & \bps        & SE-ResNet9 & 64  & 19900 & 12900 & 12600 & 72000 & 46900\tstrut\\  
    & \bpsBIG     & ResNet50   & 128 & 2300  & 1400  & 2500  & 10800 & 18400\\
    & \baselinepp & SE-ResNet9 & 64  & 2800  & 2800  & 2100  & 9300  & 13100\\
    & \baseline   & ResNet50   & 128 & 180   & 230   & 200   & 1600  & 1360\bstrut\\ 
    \midrule
    \multirow{4}{*}{\rgb}
    & \bps        & SE-ResNet9 & 64  & 13300 & 8400 & 9000 & 43000 & 37800\tstrut\\
    & \bpsBIG     & ResNet50   & 128 & 2000  & 1050 & 2200 & 6800  & 14300\\
    & \baselinepp & SE-ResNet9 & 64  & 990   &  860 & 1500 & 4600  & 8400\\    
    & \baseline   & ResNet50   & 128 & 140   &  OOM & 190  & OOM   & 1320\bstrut\\
    \bottomrule
\end{tabular}
}
\caption{\xhdr{System performance.} Average frames per second (FPS, measured as samples of experience processed per second) achieved by each system. \bps achieves a speedup of 110$\times$~ over \baseline on \depth experiments  (19,900 vs. 180~FPS) and 95$\times$ on \rgb experiments (13,300 vs. 140~FPS) on an RTX~3090 GPU.
OOM (out of memory) indicates that the RTX~2080Ti could not run \baseline with the published DD-PPO system parameters due to insufficient GPU memory.}
\label{tab:fps}
\end{table}

\begin{table}[t]
\centering
\setlength{\tabcolsep}{4pt}
\begin{tabular}{l c l c ll c cc}
    \toprule
    & \multicolumn{1}{c}{\multirow[b]{2}{*}{\bf Sensor}} &
    \multicolumn{1}{c}{\multirow[b]{2}{*}{\bf System}} &
    & \multicolumn{2}{c}{\bfseries Validation} && \multicolumn{2}{c}{\bfseries Test}\tstrut\\
    \cmidrule(lr){5-6} \cmidrule(lr){8-9}
    &&&& SPL & Success && SPL & Success\bstrut\\
    \midrule 
    \texttt{1}& 
    \multirow{2}{*}{\depth}
    & \bps &  
    &  94.4{\scriptsize$\pm$0.7} & 99.2{\scriptsize$\pm$1.4} && 91.5 & 97.3\tstrut \\
    \texttt{2}&& \baseline & 
    & 95.6{\scriptsize$\pm$0.3} & 99.9{\scriptsize$\pm$0.2} && 94.4 & 98.2\bstrut\\
    \midrule
    \texttt{3}
    &  \multirow{3}{*}{\rgb}
    & \bps & 
    & 88.4{\scriptsize$[\pm$0.9} & 97.6{\scriptsize$\pm$0.3} && 83.7 & 95.7\tstrut\\
    \texttt{4} && \bps\ {\footnotesize @ 128$\times$128} & 
    & 87.8{\scriptsize$\pm$0.7} & 97.3{\scriptsize$\pm$0.4} && 85.6 & 96.3 \\
    \texttt{5}&& \baseline & 
    & 92.9 & 99.1 && 92.0 & 97.7\bstrut\\
    \bottomrule
\end{tabular}
\caption{\xhdr{Policy performance.} SPL and Success of agents produced by \bps and \baseline. The performance of the \bps agent is within the margin of error of the \baseline agent for \depth experiments on the validation set, and within five percent on \rgb. \bps agents are trained on eight GPUs with aggregate batch size $N$=1024.}
\label{tab:spl}
\end{table}

\csubsection{End-to-End Training Speed}

\xhdr{Single-GPU performance.} On a single GPU, \bps trains agents 45$\times$ (9000 vs.~190 FPS, Tesla~V100) to 110$\times$ (19900 vs.~180 FPS, RTX~3090) faster than \baseline (Table~\ref{tab:fps}). 
The greatest speedup was achieved using the RTX 3090, which trains \depth agents at 19,900~FPS and \rgb agents at 13,300~FPS -- a 110$\times$ and 95$\times$ increase over \baseline, respectively. 
This 6000~FPS performance drop from \depth to \rgb is not caused by the more complex rendering workload, because the additional cost of fetching \rgb textures is masked by the dominant cost of geometry processing. Instead, due to memory constraints, \bps must reduce the batch size~($N$) for \rgb tasks, reducing the performance of all components (further detail in Section~\ref{sec:breakdown}). 

To assess how much of the \bps speedup is due to the SE-ResNet9 visual encoder and lower input resolution, we also compare \bpsBIG and \baseline, which have matching encoder architecture and resolution. For \depth agents training on the the RTX~3090, \bpsBIG still achieves greater than 10$\times$ performance improvement over \baseline (2,300 vs.~180~FPS), demonstrating the benefits of batch simulation even in DNN heavy workloads. \bpsBIG is only 6$\times$ faster than \baseline on the RTX~2080Ti, since the ResNet50 encoder's larger memory footprint requires batch size to be reduced from $N$=$128$ on the RTX~3090 (24~GB RAM) to $N$=$64$ on the RTX~2080Ti (11~GB RAM). Similarly, increasing DNN input resolution increases memory usage, forcing batch size to be decreased and reducing performance~(Table~\ref{tab:resfps}).

The \bps batch simulation architecture is significantly faster than the \baselinepp design that uses multiple worker processes. When training \depth agents, \bps outperforms \baselinepp by 4.5$\times$ to 7.8$\times$, with a greater speedup of 6$\times$ to 13$\times$ for \rgb agents. Since \bps and \baselinepp use the same policy DNN and input resolution, this comparison isolates the performance advantage of batch simulation and rendering against an optimized version of the multiple-worker-process-based design: \baselinepp is up to 15$\times$ faster than \baseline. The relative speedup of \bps for \rgb agents is larger because \baselinepp does not share environment assets between simulator instances.
Textures needed for \rgb rendering significantly increase the memory footprint of each simulator instance and limit \baselinepp to as few as $N$=6 workers (compared to $N$=64 for \depth agents). Conversely, \bps shares 3D assets across environments and maintains a batch size at least $N$=128 for \rgb agents.




\xhdr{Multi-GPU performance.}
\bps achieves high end-to-end throughput when running in eight-GPU configurations: up to 72,000~FPS for \depth agents on eight RTX~2080Ti. Relative to \baseline, \bps is 29$\times$ to 34$\times$ faster with eight Telsa~V100s and 45$\times$ faster with eight RTX~2080Ti. These speedups are lower than the single-GPU configurations, because \bps reduces the per-GPU batch size in eight-GPU configurations to avoid large aggregate batches that harm sample efficiency. This leads to imperfect multi-GPU scaling for \bps: for \depth agents, each RTX~2080Ti is approximately $4000$~FPS slower in an eight-GPU configuration than in a single-GPU configuration. Eight-GPU scaling for \depth is lower on the Tesla V100s (3.7$\times$) compared to the 2080Ti (5.6$\times$) because larger batch sizes are needed to utilize the large number of parallel compute units on the Tesla V100.

\csubsection{Policy Task Performance}\label{sec:spl}
\begin{figure}[t]
    \RawFloats
    \vspace{-3em}
    \centering
    \begin{minipage}[t]{.42\textwidth}
        \centering
        \includegraphics[width=0.9\textwidth]{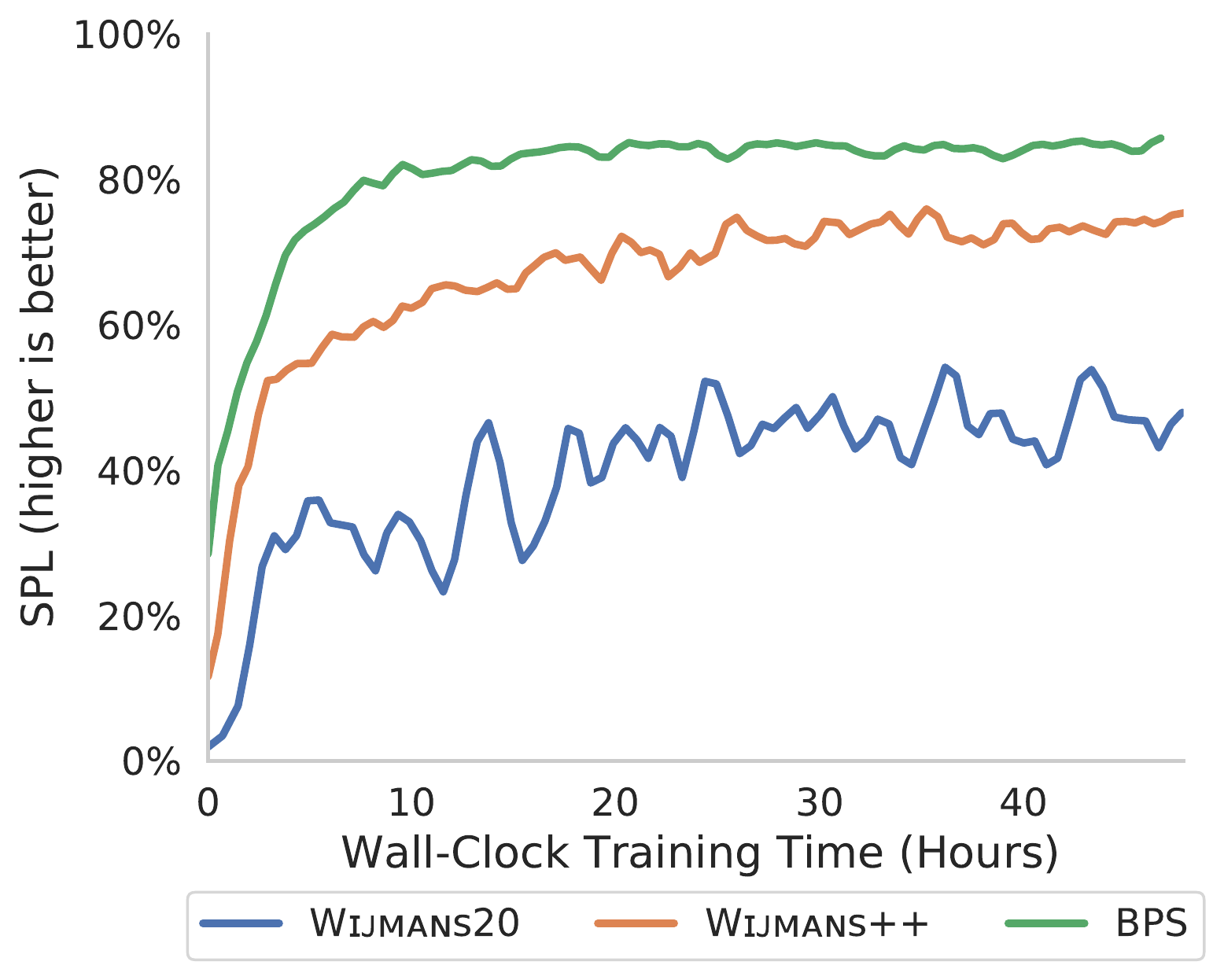}
        \vspace{0.3pt}
        \caption{SPL vs.\ wall-clock time (\rgb agents) on a RTX 3090 over 48 hours (time required to reach 2.5 billion samples with \bps). \bps exceeds $80\%$ SPL in 10 hours and achieves a significantly higher SPL than the baselines.}
        \label{fig:spltime}
    \end{minipage}
    \hspace{0.05\textwidth}
    \begin{minipage}[t]{.42\textwidth}
        \centering
        \includegraphics[width=0.9\textwidth]{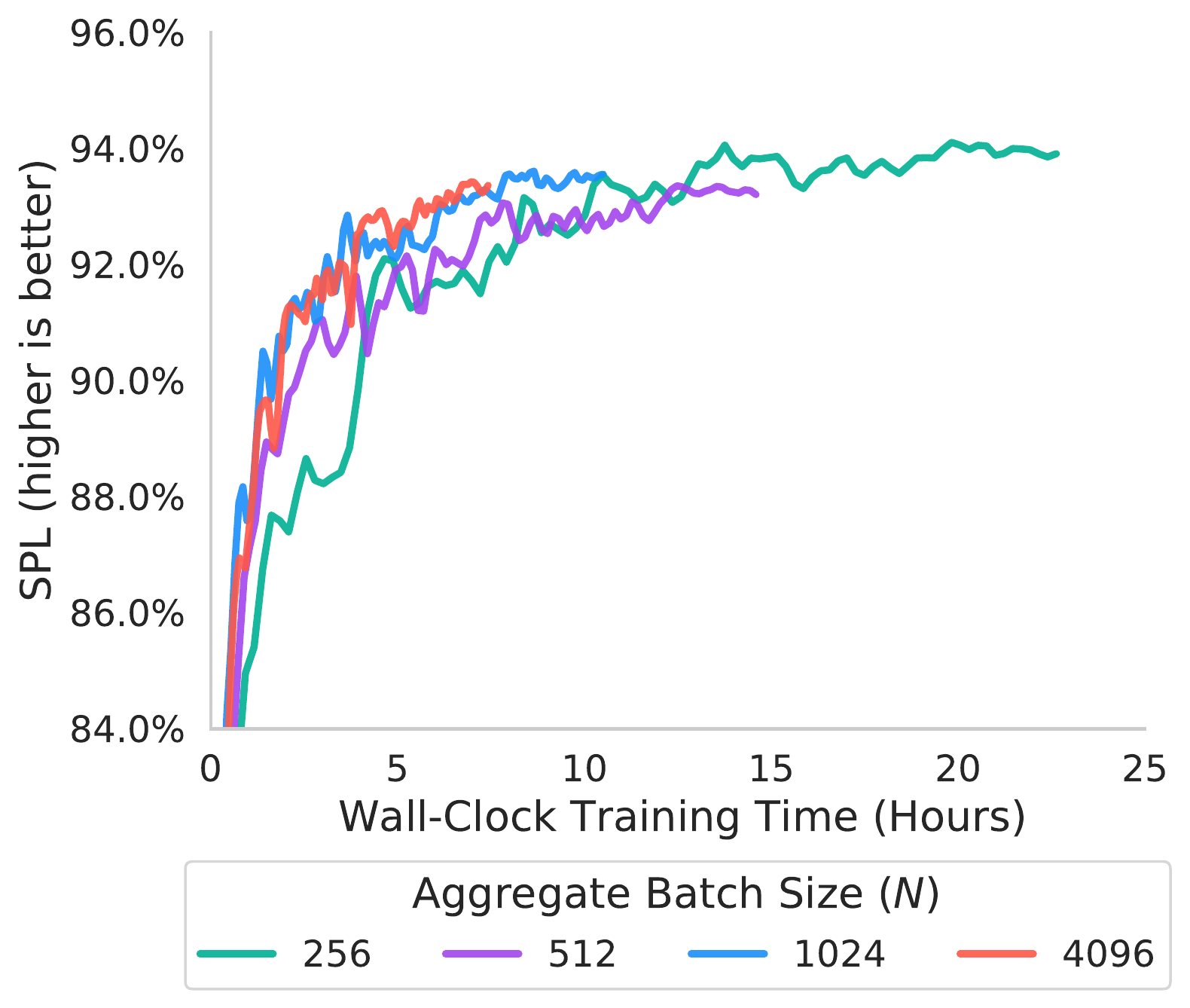}
        \vspace{0.3pt}
        \caption{SPL vs.\ wall-clock time (\bps training \depth agents over 2.5 billion samples on 8 Tesla V100s) for various batch sizes ($N$). $N$=256 finishes after 2$\times$ the wall-clock time as $N$=1024, but both achieve statistically similar SPL.}
        \label{fig:batchspl}
    \end{minipage}
\end{figure}

To understand how the system design and visual encoder architecture of \bps impact learning, we evaluate the task performance of agents trained with \bps in an eight-GPU configuration with aggregate batch size of $N$=1024.  For \depth agents, the reduction in encoder CNN depth results in a 1\% and 3\% decrease in SPL on Val and Test respectively with a negligible Success change on Val and a 0.9~Success decrease on Test (Table~\ref{tab:spl}, row~\texttt{1} vs.~\texttt{2}). 
For \rgb agents, \bps suffers a performance loss of 3.8/1.3 SPL/Success on Val and 8.3/2.0 SPL/Success on Test  (Table~\ref{tab:spl}, row~\texttt{3} vs.~\texttt{4}).  Despite this performance reduction, the \rgb agent trained by \bps would have won the 2019 Habitat challenge by 4~SPL and is only beaten by \baseline's ResNet50-based policy on Test.

\xhdr{SPL vs.~training time.} \bps significantly outperforms the baselines in terms of wall-clock training time to reach a given SPL. After 10 hours of training on a single RTX~3090, \bps reaches over $80\%$ SPL (on Val) while \baseline and \baselinepp reach only $40\%$ and $65\%$ SPL respectively (\figref{fig:spltime}). Furthermore, \bps converges within $1\%$ of peak SPL at approximately 20 hours; conversely, neither baseline reaches convergence within 48 hours. \bps converges to a lower final SPL in \figref{fig:spltime} than Table~\ref{tab:spl}, likely due to the tested single-GPU configuration differing in batch size and scene asset swapping frequency compared to the eight-GPU configuration used to produce Table~\ref{tab:spl}.

\xhdr{Effect of batch size.} The end-to-end training efficiency of \bps is dependent on batch size ($N$): larger $N$ will increase throughput and reduce wall-clock time to reach a given number of samples, but may harm sample efficiency and final task performance at convergence. We evaluate this relationship by training \depth agents with \bps across a range of $N$.
As shown in \figref{fig:batchspl}, all experiments converge within $1\%$ of the peak SPL achieved; however, $N$=256 halves total throughput compared to $N$=1024 (the setting used elsewhere in the paper for eight-GPU configurations). At the high end, $N$=4096 yields slightly worse SPL than $N$=1024 and is only 20\% faster. Larger batch sizes also require more memory for rollout storage and training, which is prohibitive for \rgb experiments that require significant GPU memory for texture assets. In terms of sample efficiency alone, \figref{fig:batchsplsteps} shows that smaller batch sizes have a slight advantage (without considering training speed).

\csubsection{Runtime Breakdown}
\label{sec:breakdown}

\begin{wrapfigure}[13]{r}{0.5\textwidth}
\begin{center}
\vspace{-35pt} 
\includegraphics[width=0.975\textwidth]{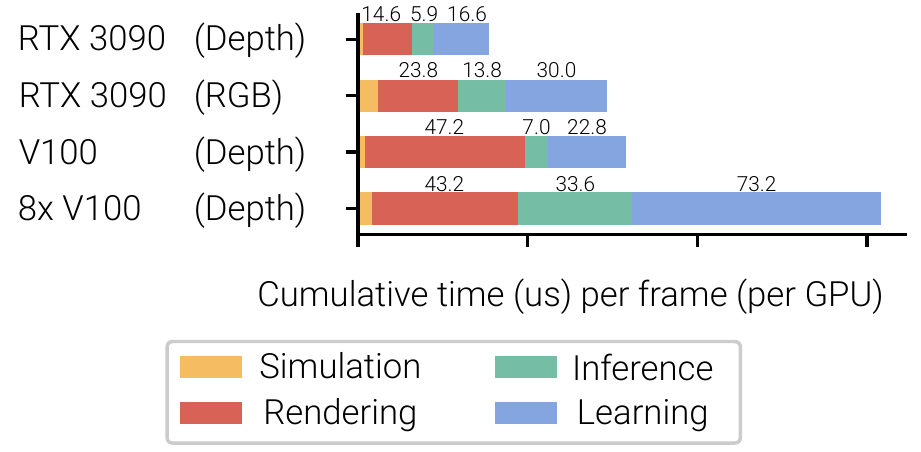}
\vspace{-5pt}
\end{center}
\caption{\bps runtime breakdown. Inference represents policy evaluation cost during rollout generation. Learning represents the total cost of policy optimization.  
}
\label{fig:breakdown}
\end{wrapfigure}

\figref{fig:breakdown} provides a breakdown of time spent in each of the main components of the \bps system ($\mu$s per frame).  
Nearly 60\% of \bps runtime on the RTX 3090 GPU (for both \depth and \rgb) is spent in DNN inference and training, even when rendering complex 3D environments and using a small, low-cost policy DNN.
This demonstrates the high degree of simulation efficiency achieved by \bps. Furthermore, the results in Table~\ref{tab:breakdown} for \bpsBIG show that, with the larger visual encoder, over 90\% of per-frame time (on \depth tasks) is spent in the DNN workload (70\% on learning).

Batch size ($N$) heavily impacts DNN performance. DNN operations for \depth~($N$=1024) are 2$\times$ faster than \rgb~($N$=256) on the RTX~3090, because \rgb must use a smaller batch size to fit texture assets in GPU memory.  The larger batch size improves GPU utilization for all system components. A similar effect is visible when comparing the single-GPU and eight-GPU V100 breakdowns. \bps reduces the per-GPU batch size from $N$=1024 to $N$=128 in eight-GPU experiments
to maintain an aggregate batch size of 1024 for sample efficiency.
Further work in policy optimization to address this learning limitation would improve multi-GPU scaling by allowing larger aggregate batch sizes.

\csection{Discussion}

We demonstrated that architecting an RL training system around the idea of batch simulation can accelerate learning in complex 3D environments by one to two orders of magnitude over prior work.
With these efficiency gains, agents can be trained with billions of simulated samples from complex environments in about a day using only a single GPU. 
We believe these fast turnaround times
stand to make RL in realistic simulated environments accessible to a broad range of researchers, increase the scale and complexity of tasks and environments that can be explored, and facilitate new studies of how much visual realism is needed to learn a given task (e.g., dynamic lighting, shadows, custom augmentations).
To facilitate such efforts, our system is available open-source at \mbox{\url{https://github.com/shacklettbp/bps-nav}}.

More generally, this work demonstrates the value of building RL systems around components that have been specifically designed for RL workloads, not repurposed from other application domains.
We believe this philosophy should be applied to other components of future RL systems, in particular to new systems for performing physics simulation in complex environments.


\subsubsection*{Acknowledgments}
This work was supported in part by NSF, DARPA, ONR YIP, ARO PECASE, Intel, and Facebook. EW is supported in part by an ARCS fellowship. We thank NVIDIA for GPU equipment donations. We also thank the Habitat team for helpful discussions and their support of this project.

\bibliography{bib/strings,bib/main}
\bibliographystyle{iclr2021_conference}

\clearpage
\appendix
\renewcommand\thefigure{A\arabic{figure}}
\renewcommand\thetable{A\arabic{table}}
\setcounter{figure}{0}
\setcounter{table}{0}
\phantomsection
\csection{Additional Results}\label{app:additional}

\begin{table}[t]
\centering
\setlength{\tabcolsep}{4pt}
\resizebox{\textwidth}{!}{
\begin{tabular}{c l r r r r r r}
    \toprule
    \multicolumn{1}{c}{\multirow[b]{2}{*}{\bf Sensor }} & \multicolumn{1}{c}{\multirow[b]{2}{*}{\bf System }} &
    \multicolumn{1}{c}{\multirow[b]{2}{*}{\bf \shortstack[c]{Agent\\Res.} }} &
    \multicolumn{1}{c}{\multirow[b]{2}{*}{\bf RTX 3090}} &
    \multicolumn{1}{c}{\multirow[b]{2}{*}{\bf RTX 2080Ti}} & \multicolumn{1}{c}{\multirow[b]{2}{*}{\bf Tesla V100}} &  \multicolumn{1}{c}{\multirow[b]{2}{*}{\bf 8$\times$2080Ti}} & \multicolumn{1}{c}{\multirow[b]{2}{*}{\bf 8$\times$V100}}\tstrut\\
    &&&&&&&\\
    \midrule 
    \multirow{4}{*}{\depth}
    & \bps        & 64  & 19900 & 12900 & 12600 & 72000 & 46900\tstrut\\  
    & \bps        & 128 & 6900  & 4880  & 5800  & 38000 & 41100\\
    & \bpsBIG     & 64  & 4800  & 2700  & 4000  & 19400 & 26500\\
    & \bpsBIG     & 128 & 2300  & 1400  & 2500  & 10800 & 18400\bstrut\\
    \midrule
    \multirow{4}{*}{\rgb}
    & \bps        & 64  & 13300 & 8400 & 9000 & 43000 & 37800\tstrut\\
    & \bps        & 128 & 6100  & 3600 & 4800 & 22300 & 31100 \\
    & \bpsBIG     & 64  & 4000  & 2100 & 3500 & 14100 & 19700\\
    & \bpsBIG     & 128 & 2000  & 1050 & 2200 & 6800  & 14300\bstrut\\
    \bottomrule
\end{tabular}
}
\caption{\xhdr{Impact of Visual Encoder Input Resolution on Performance.} Resolution has the largest impact on performance when increased memory usage forces \bps' batch size to be decreased. For example, on a single Tesla~V100, \bps' \depth performance drops by 2.2$\times$ after increasing the resolution, because batch size decreases from $N$=1024 to $N$=512. Conversely, the eight-GPU Tesla~V100 results only show a 12\% decrease in performance, since batch size is fixed at $N$=128. Experiments with 128$\times$128 pixel resolution are rendered at 256$\times$256 and downsampled.}
\label{tab:resfps}
\end{table}

\begin{table}[t]
\centering
\setlength{\tabcolsep}{4pt}
\begin{tabular}{l l c c rrr}
    \toprule
     \bf \shortstack[c]{Sensor} & \bf System & \bf CNN && \bf \shortstack[c]{Simulation + \\Rendering} & \bf Inference & \bf Learning\tstrut\bstrut\\
    \midrule
    \multirow{4}{*}{\depth} & \bps & SE-ResNet9 && 16.1 & 5.9 & 16.6\tstrut\\
    & \bpsBIG & ResNet50 && 26.9 & 99.3 & 311.3 \\
    & \baselinepp & SE-ResNet9 && 270.9 & 78.8 & 42.8 \\
    & \baseline & ResNet50 && 1901.3 & 3968.6 & 1534.5\bstrut\\
    \midrule
    \multirow{4}{*}{\rgb} & \bps & SE-ResNet9 && 29.6 & 13.8 & 30.0\tstrut\\
    & \bpsBIG & ResNet50 && 40.3 & 110.2 & 333.4  \\
    & \baselinepp & SE-ResNet9 && 520.3 & 389.5 & 169.3 \\
    & \baseline & ResNet50 && 1911.1 & 4027.5 & 1587.5\bstrut\\
    \bottomrule
\end{tabular}
\caption{\xhdr{Runtime breakdown across systems.} Microseconds per frame for each RL component on a RTX 3090. SE-ResNet9 uses an input resolution of 64x64, while ResNet50 uses an input resolution of 128x128. Note the large amount of time spent by \baseline on policy inference, caused by GPU memory constraints that force a small number of rollouts per iteration. \bpsBIG's performance is dominated by the DNN workload due to the large ResNet50 visual encoder.}
\label{tab:breakdown}
\end{table}

\begin{table}[t]
\centering
\begin{tabular}{l r r r}
\toprule
\bf Task & \bf FPS & \bf Training Score & \bf Validation Score\tstrut\bstrut\\
\midrule
Explore & 25300 & 6.42 & 5.61\tstrut\\ 
Flee & 24700 & 4.27 & 3.65\bstrut\\ 
\bottomrule
\end{tabular}
\caption{\xhdr{Task and FPS results for Flee and Explore tasks} with \depth agents (on a RTX 3090), where the Training / Validation Score is measured in meters for the Flee task and number of cells visited on the navigation mesh for the Explore task. These tasks achieve higher throughput than \pointnavfull due to the lower complexity AI2-THOR meshes used. The relatively low scores are a result of the small spatial size of the AI2-THOR assets.}
\label{tab:fleeexplore}
\end{table}
\csubsection{Flee and Explore Tasks on AI2-THOR Dataset}
To demonstrate batch simulation and rendering on additional tasks besides \pointnavfull, \bps also supports the Flee (find the farthest valid location from a given point) and Explore (visit as much of an area as possible) tasks. We evaluate \bps's performance on these tasks on the AI2-THOR \citep{ai2thor} dataset to additionally show how batch rendering performs on assets with less geometric complexity than the scanned geometry in Gibson and \matterport.

Table~\ref{tab:fleeexplore} shows the learned task performance and end-to-end training speed of \bps on these two tasks for \depth-sensor-driven agents. For both tasks, \bps outperforms its results on \pointnavfull by around 5000 frames per second, largely due to the significantly reduced geometric complexity of the AI2-THOR dataset versus Gibson. Additionally, the Explore task slightly outperforms the Flee task by 600 FPS on average due to a simpler simulation workload, because no geodesic distance computation is necessary.

\csubsection{Standalone Batch Renderer Performance}\label{app:renderer}
\begin{figure}[t]
    \RawFloats
    \centering
    \vspace{2em}
    \begin{minipage}[t]{.45\textwidth}
        \centering
        \includegraphics[width=0.9\textwidth]{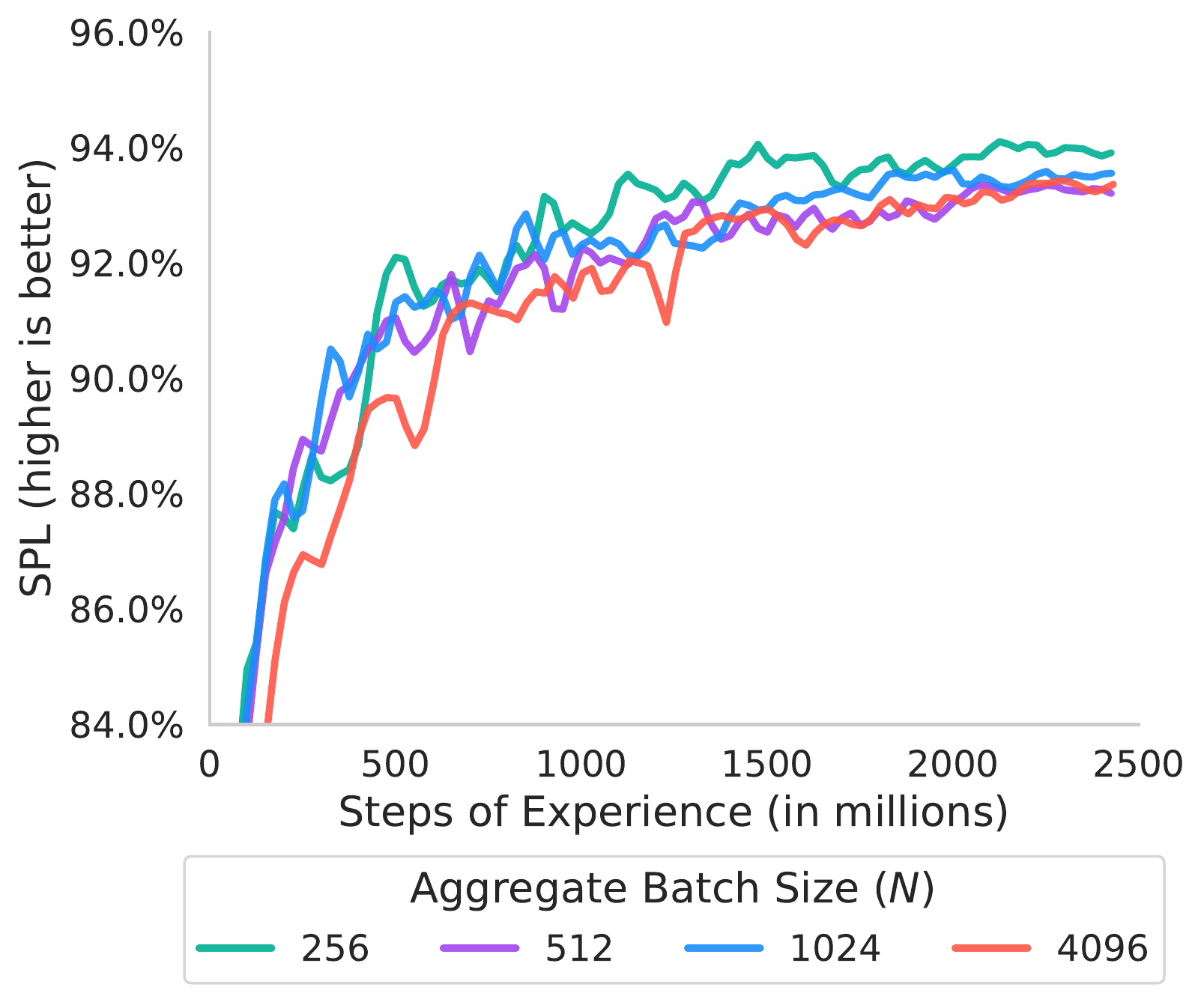}
        \vspace{0.1pt}
        \caption{\bps's validation set SPL for \depth vs. number of training samples across a range of batch sizes. This graph shows that sample efficiency slightly decreases with larger batch sizes (with the exception of $N$=512 vs. $N$=1024, where $N$=1024 exhibits better validation score). Ultimately, the difference in converged performance is less than 1\% SPL between different batch sizes. Although $N$=256 converges the fastest in terms of training samples needed, \figref{fig:batchspl} shows that $N$=256 performs poorly in terms of SPL achieved per unit of training time. }
        \label{fig:batchsplsteps}
    \end{minipage}
    \hspace{0.05\textwidth}
    \begin{minipage}[t]{.45\textwidth}
        \centering
        \includegraphics[width=0.9\textwidth]{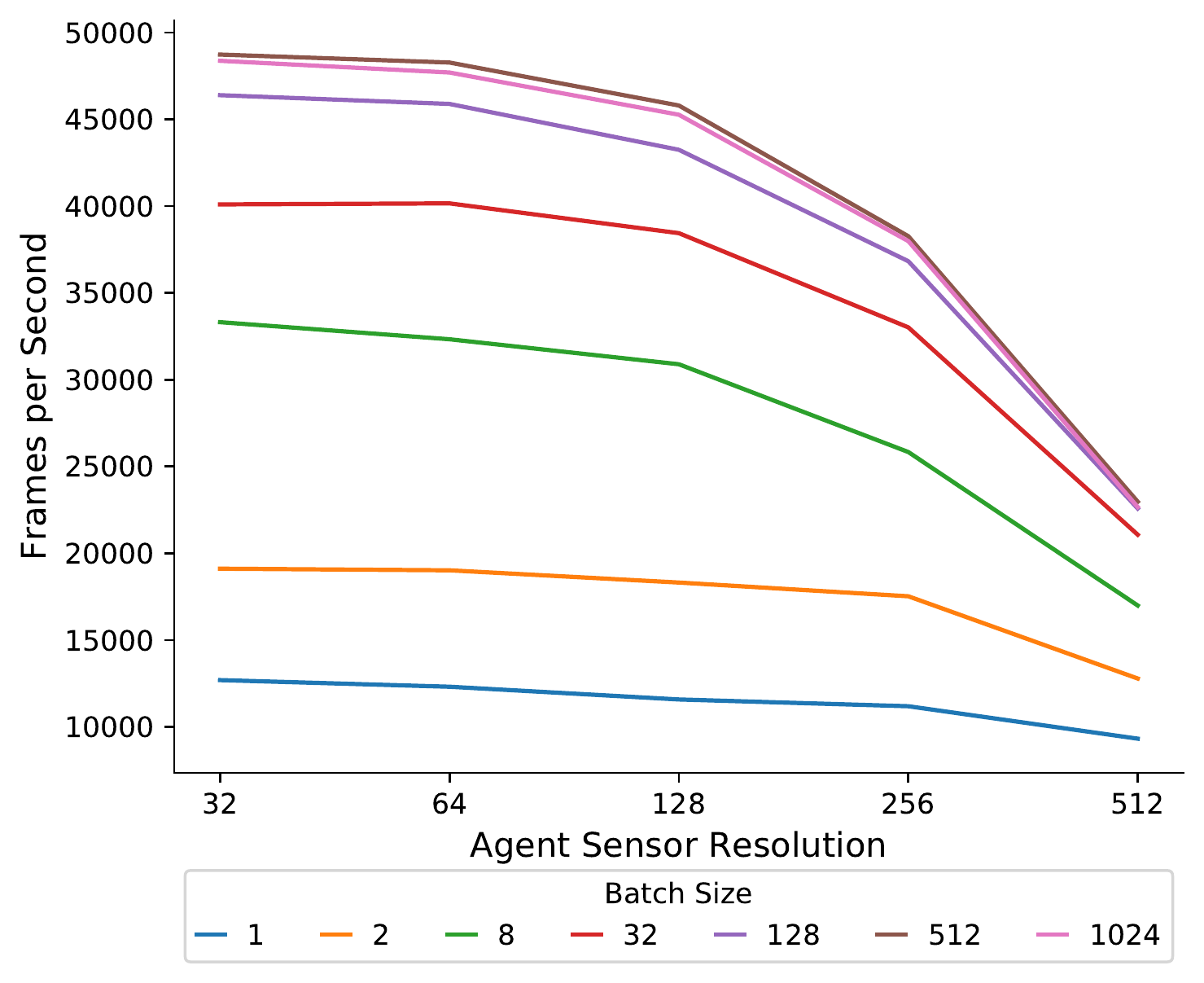}
        \vspace{0.1pt}
        \caption{Frames per second achieved by the standalone renderer on a RTX 3090 across a range of resolutions and batch sizes for a \rgb sensor on the Gibson dataset. Performance saturates at a batch size of 512. For lower batch sizes, increasing resolution has a minimal performance impact, because the GPU still isn't fully utilized. As resolution increases with larger batches, the relative decrease in performance from higher resolution increases.}
        \label{fig:renderspeed}
    \end{minipage}
\end{figure}

To evaluate the absolute performance of \bps's batch renderer independently from other components of the system, \figref{fig:renderspeed} shows the performance of the standalone renderer on the ``Stokes'' scene from the Gibson dataset using a set of camera positions taken from a training run. A batch size of 512 achieves a 3.7x performance increase over a batch size of 1, which emphasizes the fact that much of the end to end speedup provided by batch rendering comes from the performance benefits of larger inference and training batches made possible by the batch renderer's 3D asset sharing.

\figref{fig:renderspeed} also demonstrates that the batch renderer can maintain extremely high performance (approximately 23,000 FPS) at much higher resolutions than used in the RL tasks presented in this work. While this may be useful for tasks requiring higher resolution inputs, considerable advancements would need to be made in DNN performance to handle these high resolution frames at a comparable framerate to the renderer.

\csubsection{Lamb Optimizer Ablation Study}
\begin{figure}[t]
    \RawFloats
        \centering
        \includegraphics[width=0.6\textwidth]{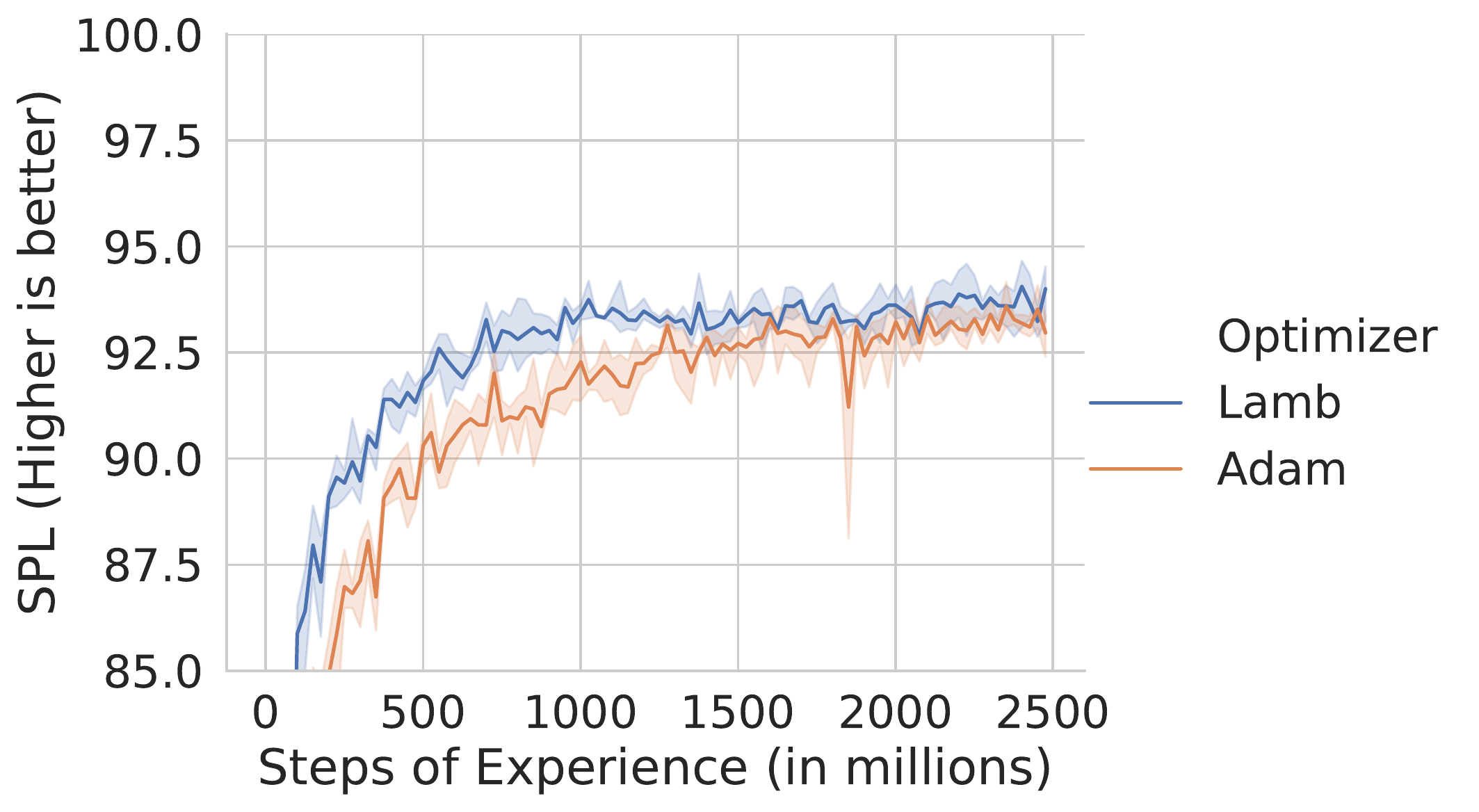}
        \vspace{0.1pt}
        \caption{The effect of the Lamb optimizer versus the baseline Adam optimizer on sample efficiency while training a \depth sensor driven agent. Lamb maintains a consistent lead in terms of SPL throughout training, especially in the first half of training.}
        \label{fig:lamb}
\end{figure}

To demonstrate the benefit provided by the Lamb optimizer with regard to sample efficiency, \figref{fig:lamb} shows a comparison between the Lamb optimizer used by \bps and the Adam optimizer used by \baseline and \baselinepp. The training setup for these two optimizers is identical, with the exception of the removal of learning rate scaling for Adam, as this causes training to diverge. The benefits of Lamb are most pronounced early in training, allowing Lamb to reach within 0.7\% SPL of convergence after just 1 billion samples of experience (while Adam trails Lamb by 1.5\% at the same point). As training progresses, the difference shrinks as Adam slowly converges for a final difference of 0.6\% SPL after 2.5 billion frames.
 
\csection{Experiment and Training Additional Details}
\label{experiment_appendix}

\xhdr{Complete \pointnavfull description.} We train and evaluate agents via the same procedure as~\citet{ddppo}.  Specifically, agents are trained for \pointnav~\citep{anderson2018evaluation} where the agent is tasked with navigating to a point specified relative to its initial location.
Agents are equipped with a \gpscompass sensor (providing the agent with its position and orientation relative to the starting position) and either a \depth sensor or \rgb camera. The agent has access to 4 low-level actions, \texttt{forward} (0.25m), \texttt{turn\_left}(10$^\circ$), \texttt{turn\_right}(10$^\circ$), and \texttt{stop}.

Agents are evaluated on the Gibson dataset~\citep{xia2018gibson}. We use two metrics to evaluate the agents: Success, whether or not the agent called \texttt{stop} within 0.2m of the goal, and SPL~\citep{anderson2018evaluation}, a measure of both Success and efficiency of the agent's path. During evaluation, the agent does not have access to reward. 

\xhdr{Half-precision inference and mixed-precision training.} We perform inference in half precision for all components except the action distribution. We train in mixed precision~\citep{jia2018highly}, utilizing the Apex library in O2 mode.  We use half precision for all computations except the action distribution and losses. Additionally, The optimizer still utilizes single precision for all computations and applies gradients to a single-precision copy of the weights.

\xhdr{Training hyper-parameters} Our hyper-parameters for eight-GPU runs are given in~\reftab{tab:hparams}.  We additionally employ a gradual learning rate decay where we decay the learning rate from its scaled value back to the base value over the first half of training.  We use a cosine schedule.

We find it necessary to set $\rho{=}1.0$ for the bias parameters, fixup parameters, and layer-norm parameters of the network, making the optimizer for these parameters equivalent to AdamW~\citep{kingma2016adam,loshchilov2018decoupled}.
We also use L2 weight-decay both to add back regularization lost by removing normalization layers and to stabilize Lamb; we use $\lambda{=}10^{-2}$.

We find one epoch of PPO with two mini-batches to be sufficient (instead of two epochs with two mini-batches), thus effectively doubling the learning speed.  We also evaluated one mini-batch, but found two to be beneficial while also having little penalty on overall training speed.

\begin{table}
    \centering
    \begin{tabular}{l c}
    \toprule
    \multicolumn{2}{c}{PPO Parameters} \\ \\
    PPO Epochs & 1 \\
    PPO Mini-Batches & 2 \\
    PPO Clip & 0.2 \\
    Clipped value loss & No \\
    Per mini-batch advantage normalization & No \\
    $\gamma$ & 0.99 \\
    GAE-$\lambda$~\citep{schulman2016high} & 0.95 \\
    Learning rate & $5.0\times 10^{-4}$ \depth,  $2.5\times 10^{-4}$ \rgb\\
    Learning rate scaling & $\sqrt{\frac{B}{B_{\text{base}}}}$ \\
    $B_{\text{base}}$ & 256 \\
    Max gradient norm & $1.0$ \\
    Weight decay & $0.01$ \\
    Lamb $\rho$ & $0.01$ \\
    \midrule 
    \multicolumn{2}{c}{Per GPU parameters} \\ \\
    Number of unique scenes ($K$) & 4 \\
    Simulation batch size/Number of Environments ($N$) & 128 \\
    Rollout length ($L$) & 32 \\
    \bottomrule
    \end{tabular}
    \caption{Hyper-parameters used for \bps training on 8 GPUs.}
    \label{tab:hparams}
\end{table}

\csection{Benchmarking Additional Details}
\label{hwappendix}

\xhdr{Pretrained benchmarking.} A pretrained DNN is used when benchmarking to avoid frequent environment resets at the start of training.

\xhdr{Benchmarking hyper-parameters.} \reftab{tab:benchparams} shows the setting for hyper-parameters that impact system throughput.

\begin{table}[t]
\caption{System configuration parameters for \reftab{tab:fps}. $^*$ indicates 4 mini batches per epoch instead of 2.}
\label{tab:benchparams}
\begin{center}
\setlength{\tabcolsep}{4pt}
\resizebox{0.975\textwidth}{!}{
\begin{tabular}{c lcclc rrc rrc r}
    \toprule
    \multicolumn{1}{c}{\multirow[b]{2}{*}{\bf Sensor }} & \multicolumn{1}{c}{\multirow[b]{2}{*}{\bf System }} & \multicolumn{1}{c}{\multirow[b]{2}{*}{\bf CNN }} & \multicolumn{1}{c}{\multirow[b]{2}{*}{\bf Resolution }} &&& \multicolumn{2}{c}{\bfseries Tesla V100} && \multicolumn{2}{c}{\bfseries RTX 2080Ti} && \multicolumn{1}{c}{\bfseries RTX 3090}\tstrut \\
    \cmidrule(lr){7-8} \cmidrule(lr){10-11} \cmidrule(lr){13-13}
    & && & && 1 GPU & 8 GPUs && 1 GPU & 8 GPUs && \multicolumn{1}{c}{1 GPU}\bstrut \\
    \midrule 
    \multirow{18}{*}{\depth} 
    & \multirow{3}{*}{\bps} & \multirow{3}{*}{SE-ResNet9} & \multirow{3}{*}{64} & PPO Epochs && \multicolumn{7}{c}{1} \\\bstrut
    &&& &  Rollout length ($L$) &&  \multicolumn{7}{c}{32} \\
    &&& & Number of Environments ($N$) && 1024 & 128 && 512 & 128 && 1024 \\\tstrut
    & \multirow{3}{*}{\bps} & \multirow{3}{*}{SE-ResNet9} & \multirow{3}{*}{128} & PPO Epochs && \multicolumn{7}{c}{1} \\\bstrut
    &&& &  Rollout length ($L$) &&  \multicolumn{7}{c}{32} \\
    &&& & Number of Environments ($N$) && 512 & 128 && 128 & 128 && 512 \\\tstrut
    & \multirow{3}{*}{\bpsBIG} & \multirow{3}{*}{ResNet50} & \multirow{3}{*}{64} & PPO Epochs && \multicolumn{7}{c}{1} \\\bstrut
    &&& &  Rollout length ($L$) &&  \multicolumn{7}{c}{32} \\
    &&& & Number of Environments ($N$) && 512 & 128 && 256 & 128 && 512 \\\tstrut
    & \multirow{3}{*}{\bpsBIG} & \multirow{3}{*}{ResNet50} & \multirow{3}{*}{128} & PPO Epochs && \multicolumn{7}{c}{1} \\\bstrut
    &&& &  Rollout length ($L$) &&  \multicolumn{7}{c}{32} \\
    &&& & Number of Environments ($N$) && 256 & 128 && 64 & 64 && 128 \\\tstrut
    & \multirow{3}{*}{\baselinepp} & \multirow{3}{*}{SE-ResNet9} & \multirow{3}{*}{64} & PPO Epochs && \multicolumn{7}{c}{1} \\\bstrut
    &&& &  Rollout length ($L$) &&  \multicolumn{7}{c}{32} \\
    &&& & Number of Environments ($N$) && \multicolumn{7}{c}{64} \\\tstrut
    & \multirow{3}{*}{\baseline} & \multirow{3}{*}{ResNet50} & \multirow{3}{*}{128} & PPO Epochs && \multicolumn{7}{c}{2} \\\bstrut
    &&& &  Rollout length ($L$) &&  \multicolumn{7}{c}{128} \\
    &&& & Number of Environments ($N$) && \multicolumn{7}{c}{4} \\
    \midrule
    \multirow{18}{*}{\rgb} 
    & \multirow{3}{*}{\bps} & \multirow{3}{*}{SE-ResNet9} & \multirow{3}{*}{64} & PPO Epochs && \multicolumn{7}{c}{1} \\\bstrut
    &&& &  Rollout length ($L$) &&  \multicolumn{7}{c}{32} \\
    &&& & Number of Environments ($N$) && 512 & 128 && 128 & 128 && 256 \\\tstrut
    & \multirow{3}{*}{\bps} & \multirow{3}{*}{SE-ResNet9} & \multirow{3}{*}{128} & PPO Epochs && \multicolumn{7}{c}{1} \\\bstrut
    &&& &  Rollout length ($L$) &&  \multicolumn{7}{c}{32} \\
    &&& & Number of Environments ($N$) && 256 & 128 && 64$^*$ & 64$^*$ && 256 \\\tstrut
    & \multirow{3}{*}{\bpsBIG} & \multirow{3}{*}{ResNet50} & \multirow{3}{*}{64} & PPO Epochs && \multicolumn{7}{c}{1} \\\bstrut
    &&& &  Rollout length ($L$) &&  \multicolumn{7}{c}{32} \\
    &&& & Number of Environments ($N$) && 256 & 128 && 64 & 64 && 256 \\\tstrut
    & \multirow{3}{*}{\bpsBIG} & \multirow{3}{*}{ResNet50} & \multirow{3}{*}{128} & PPO Epochs && \multicolumn{7}{c}{1} \\\bstrut
    &&& &  Rollout length ($L$) &&  \multicolumn{7}{c}{32} \\
    &&& & Number of Environments ($N$) && 128 & 128 && 32$^*$ & 32$^*$ && 64 \\\tstrut
    & \multirow{3}{*}{\baselinepp} & \multirow{3}{*}{SE-ResNet9} & \multirow{3}{*}{64} & PPO Epochs && \multicolumn{7}{c}{1} \\\bstrut
    &&& &  Rollout length ($L$) &&  \multicolumn{7}{c}{32} \\
    &&& & Number of Environments ($N$) && 20 & 20 && 6 & 6 && 16 \\\tstrut
    & \multirow{3}{*}{\baseline} & \multirow{3}{*}{ResNet50} & \multirow{3}{*}{128} & PPO Epochs && \multicolumn{7}{c}{2} \\\bstrut
    &&& &  Rollout length ($L$) &&  \multicolumn{7}{c}{128} \\
    &&& & Number of Environments ($N$) && \multicolumn{7}{c}{4} \\
    \bottomrule
\end{tabular}
}
\end{center}
\end{table}

\xhdr{GPU details} We report FPS results on three models of NVIDIA GPUs: Tesla V100, GeForce RTX 2080 TI, and GeForce RTX 3090. We demonstrate scaling to multiple GPUs with eight GPU configurations for all but the RTX 3090. Single GPU and eight GPU results are benchmarked on the same machines; however single GPU configurations are limited to 12 cores and 64~GB of RAM as this is a reasonable configuration for a single GPU workstation.

\xhdr{CPU details}.  Each GPU configuration also uses different CPU configurations based on hardware access. Tesla V100 benchmarking was done with 2x Intel Xeon E5-2698 v4 (a DGX-1 station). RTX 2080 TI benchmarking was done with 2x Intel Xeon Gold 6226. RTX 3090 benchmarking was done with with 1x Intel i7-5820k.  On all CPUs, we  disable Hardware P-State (HWP) (where applicable) and put software P-State in performance mode. Our CPU load on simulation worker cores is inherently sporadic and we find that certain CPUs are unable to change clock frequencies fast enough to not incur a considerable performance penalty when allowed to enter a power saving state.

\end{document}

%% file: standalone/definitions.tex
\newcommand{\myapprox}{{\raise.17ex\hbox{$\scriptstyle\sim$}}}
\newcommand{\xhdr}[1]{\vspace{0pt}\noindent\textbf{#1}\xspace}

\newcommand{\reftab}[1]{Table~\ref{#1}}

\makeatletter
\newcommand\footnoteref[1]{\protected@xdef\@thefnmark{\ref{#1}}\@footnotemark}
\makeatother

\newcommand{\matterport}{Matterport3D\xspace}

\newcommand{\rgb}{\texttt{RGB}\xspace}
\newcommand{\depth}{\texttt{Depth}\xspace}

\newcommand{\pointnav}{\texttt{PointGoalNav}\xspace}
\newcommand{\pointnavfull}{PointGoal navigation\xspace}
\newcommand{\compassgps}{GPS+Compass\xspace}
\newcommand{\gpscompass}{\compassgps}

\newcommand{\baseline}{\textsc{w{\fontsize{8.5}{10}\selectfont
ijmans}\raisebox{-.025ex}{\fontsize{8.2}{10}\selectfont 20}}\xspace}
\newcommand{\baselinepp}{\textsc{w{\fontsize{8.5}{10}\selectfont
ijmans}\nolinebreak[4]\hspace{0.01em}\raisebox{0.1ex}{\footnotesize\bf ++}}\xspace}
\newcommand{\bps}{\textsc{bps}\xspace}
\newcommand{\bpsBIG}{\textsc{bps-r\raisebox{-.025ex}{\fontsize{8.2}{10}\selectfont 50}}\xspace}

%% file: standalone/space_saver.tex
\belowdisplayskip \abovedisplayskip

\newcommand{\csection}[1]{
    \vspace{-0.04in}
    \section{#1}
    \vspace{-0.04in}
}

\newcommand{\csubsection}[1]{
    \vspace{-0.04in}
    \subsection{#1}
    \vspace{-0.04in}
}

%% file: sections/main/related-work.tex
\csection{Related Work}

\xhdr{Systems for high-performance RL.} Existing systems for high-performance RL have primarily focused on improving the efficiency of DNN components of the workload (policy inference and optimization) and use a simulator designed for efficient \emph{single} agent simulation as a black box. For example, Impala and Ape-X used multiple worker processes to asynchronously collect experience for a centralized learner~\citep{espeholt2018impala,horgan2018distributed}. SEED RL and Sample Factory built upon this idea and introduced inference workers that centralize network inference, thereby allowing it to be accelerated by GPUs or TPUs~\citep{espeholt2020seed, petrenko2020sample}. DD-PPO proposed a synchronous distributed system for similar purposes \citep{ddppo}. A number of efficient implementations 
of these ideas have been proposed as part of RL frameworks or in other deep learning libraries~\citep{liang2018rllib,stooke2019rlpyt,kttler2019torchbeast}.    

We extend the idea of centralizing inference and learning to simulation 
by cracking open the simulator black box and designing a new simulation architecture for RL workloads.
Our large-batch simulator 
is a drop-in replacement for large numbers of (non-batched) simulation workers,
making it synergistic with existing asynchronous and synchronous distributed training schemes.  
It reduces the number of processes and communication overhead needed for asynchronous methods and eliminates separate simulation worker processes altogether for synchronous methods.
We demonstrate this by combining our system with DD-PPO~\citep{ddppo}.

Concurrently with our work, CuLE, a GPU-accelerated reimplementation of the Atari Learning Environment (ALE), demonstrates the benefits of centralized batch simulation~\citep{cule}. While both our work and CuLE enable wide-batch execution of their respective simulation workloads, our focus is on high-performance batch rendering of complex 3D environments.  This involves optimizations (GPU-driven pipelined geometry culling, 3D asset sharing, and asynchronous data transfer) not addressed by CuLE due to the simplicity of rendering Atari-like environments. Additionally, like CuLE, we observe that the large training batches produced by batch simulation reduce RL sample efficiency. Our work goes further and leverages large-batch optimization techniques from the supervised learning literature to mitigate the loss of sample efficiency without shrinking batch size.

\xhdr{Large mini-batch optimization.} 
A consequence of large batch simulation is that more experience is collected between gradient updates.  This provides the opportunity to accelerate learning via large mini-batch optimization.
In supervised learning, using large mini-batches during optimization typically decreases the generalization performance of models~\citep{keskar2016large}. 
\citet{goyal2017accurate} demonstrated that model performance can be improved by scaling the learning rate proportionally with 
the batch size and ``warming-up'' the learning rate at the start of training.
\citet{you2017scaling} proposed an optimizer modification, LARS, that adaptively scales the learning rate 
at each layer, and applied it to SGD to improve generalization further. 
In reinforcement learning and natural language processing, the Adam optimizer~\citep{kingma2016adam} is often used instead of SGD.
Lamb~\citep{you2020large} combines LARS~\citep{you2017scaling} with Adam~\citep{kingma2016adam}. 
We do not find that large mini-batch optimization harms generalization in reinforcement learning, but 
we do find it decreases sample efficiency. 
We adapt the techniques proposed above -- learning rate scaling~\citep{you2017scaling} and the Lamb optimizer~\citep{you2020large} -- 
to improve sample efficiency.

\xhdr{Simulators for machine learning.}
Platforms for simulating realistic environments for model training fall into two broad categories:
those built on top of pre-existing game engines~\citep{ai2thor,dosovitskiy2017carla,lee2019ikea,tdw,james2019rlbench}, 
and those built from scratch using open-source 3D graphics and physics libraries~\citep{savva2017minos,habitat19iccv,xia2018gibson,xia2020interactive,Xiang2020,zeng2020transporter}.
While improving simulator performance has been a focus of this line of work, 
it has been evaluated in a narrow sense (\ie frame rate benchmarks for predetermined agent trajectories),
not accounting for the overall performance of end-to-end RL training.
We instead take a holistic approach to co-design rendering and simulation modules and their interfaces to the RL training system,
obtaining significant gains in end-to-end throughput over the state of the art.


%% file: iclr2021_conference.bbl
\begin{thebibliography}{47}
\providecommand{\natexlab}[1]{#1}
\providecommand{\url}[1]{\texttt{#1}}
\expandafter\ifx\csname urlstyle\endcsname\relax
  \providecommand{\doi}[1]{doi: #1}\else
  \providecommand{\doi}{doi: \begingroup \urlstyle{rm}\Url}\fi

\bibitem[Akenine-M{\"o}ller et~al.(2018)Akenine-M{\"o}ller, Haines, and
  Hoffman]{akenine2019real}
Tomas Akenine-M{\"o}ller, Eric Haines, and Naty Hoffman.
\newblock \emph{Real-time rendering}.
\newblock CRC Press, 2018.

\bibitem[Anderson et~al.(2018)Anderson, Chang, Chaplot, Dosovitskiy, Gupta,
  Koltun, Kosecka, Malik, Mottaghi, Savva, et~al.]{anderson2018evaluation}
Peter Anderson, Angel Chang, Devendra~Singh Chaplot, Alexey Dosovitskiy,
  Saurabh Gupta, Vladlen Koltun, Jana Kosecka, Jitendra Malik, Roozbeh
  Mottaghi, Manolis Savva, et~al.
\newblock On evaluation of embodied navigation agents.
\newblock \emph{arXiv:1807.06757}, 2018.

\bibitem[Beattie et~al.(2016)Beattie, Leibo, Teplyashin, Ward, Wainwright,
  K{\"u}ttler, Lefrancq, Green, Vald{\'e}s, Sadik, et~al.]{beattie2016deepmind}
Charles Beattie, Joel~Z Leibo, Denis Teplyashin, Tom Ward, Marcus Wainwright,
  Heinrich K{\"u}ttler, Andrew Lefrancq, Simon Green, V{\'\i}ctor Vald{\'e}s,
  Amir Sadik, et~al.
\newblock Deepmind lab.
\newblock \emph{arXiv:1612.03801}, 2016.

\bibitem[Bellemare et~al.(2013)Bellemare, Naddaf, Veness, and
  Bowling]{bellemare2013arcade}
Marc~G Bellemare, Yavar Naddaf, Joel Veness, and Michael Bowling.
\newblock The arcade learning environment: An evaluation platform for general
  agents.
\newblock \emph{Journal of Artificial Intelligence Research}, 47, 2013.

\bibitem[{Carnegie Mellon University}(2019)]{locobot}
{Carnegie Mellon University}.
\newblock Locobot: An open source low cost robot.
\newblock \url{https://locobot-website.netlify.com/}, 2019.

\bibitem[Chang et~al.(2017)Chang, Dai, Funkhouser, Halber, Niessner, Savva,
  Song, Zeng, and Zhang]{mp3d}
Angel Chang, Angela Dai, Thomas Funkhouser, Maciej Halber, Matthias Niessner,
  Manolis Savva, Shuran Song, Andy Zeng, and Yinda Zhang.
\newblock {Matterport3D}: Learning from {RGB-D} data in indoor environments.
\newblock In \emph{International Conference on 3D Vision (3DV)}, 2017.
\newblock {MatterPort3D} dataset license available at:
  \url{http://kaldir.vc.in.tum.de/matterport/MP_TOS.pdf}.

\bibitem[Dalton et~al.(2020)Dalton, Frosio, and Garland]{cule}
Steven Dalton, Iuri Frosio, and Michael Garland.
\newblock Accelerating reinforcement learning through gpu atari emulation.
\newblock \emph{NeurIPS}, 2020.

\bibitem[Dosovitskiy et~al.(2017)Dosovitskiy, Ros, Codevilla, Lopez, and
  Koltun]{dosovitskiy2017carla}
Alexey Dosovitskiy, German Ros, Felipe Codevilla, Antonio Lopez, and Vladlen
  Koltun.
\newblock {CARLA}: {An} open urban driving simulator.
\newblock In \emph{Proceedings of the 1st Annual Conference on Robot Learning},
  pp.\  1--16, 2017.

\bibitem[Espeholt et~al.(2018)Espeholt, Soyer, Munos, Simonyan, Mnih, Ward,
  Doron, Firoiu, Harley, Dunning, et~al.]{espeholt2018impala}
Lasse Espeholt, Hubert Soyer, Remi Munos, Karen Simonyan, Volodymir Mnih, Tom
  Ward, Yotam Doron, Vlad Firoiu, Tim Harley, Iain Dunning, et~al.
\newblock Impala: Scalable distributed deep-rl with importance weighted
  actor-learner architectures.
\newblock In \emph{Proceedings of the International Conference on Machine
  Learning (ICML)}, 2018.

\bibitem[Espeholt et~al.(2020)Espeholt, Marinier, Stanczyk, Wang, and
  Michalski]{espeholt2020seed}
Lasse Espeholt, Rapha{\"e}l Marinier, Piotr Stanczyk, Ke~Wang, and Marcin
  Michalski.
\newblock Seed rl: Scalable and efficient deep-rl with accelerated central
  inference.
\newblock In \emph{Proceedings of the International Conference on Learning
  Representations (ICLR)}, 2020.

\bibitem[Gan et~al.(2020)Gan, Schwartz, Alter, Schrimpf, Traer, De~Freitas,
  Kubilius, Bhandwaldar, Haber, Sano, et~al.]{tdw}
Chuang Gan, Jeremy Schwartz, Seth Alter, Martin Schrimpf, James Traer, Julian
  De~Freitas, Jonas Kubilius, Abhishek Bhandwaldar, Nick Haber, Megumi Sano,
  et~al.
\newblock Threedworld: A platform for interactive multi-modal physical
  simulation.
\newblock \emph{arXiv:2007.04954}, 2020.

\bibitem[Goyal et~al.(2017)Goyal, Doll{\'{a}}r, Girshick, Noordhuis,
  Wesolowski, Kyrola, Tulloch, Jia, and He]{goyal2017accurate}
Priya Goyal, Piotr Doll{\'{a}}r, Ross~B. Girshick, Pieter Noordhuis, Lukasz
  Wesolowski, Aapo Kyrola, Andrew Tulloch, Yangqing Jia, and Kaiming He.
\newblock Accurate, large minibatch {SGD}: Training {ImageNet} in 1 hour.
\newblock \emph{arXiv:1706.02677}, 2017.

\bibitem[He et~al.(2016)He, Zhang, Ren, and Sun]{he2016resnet}
Kaiming He, Xiangyu Zhang, Shaoqing Ren, and Jian Sun.
\newblock Deep residual learning for image recognition.
\newblock In \emph{Proceedings of IEEE Conference on Computer Vision and
  Pattern Recognition (CVPR)}, 2016.

\bibitem[Hochreiter \& Schmidhuber(1997)Hochreiter and
  Schmidhuber]{hochreiter97lstm}
Sepp Hochreiter and J{\"{u}}rgen Schmidhuber.
\newblock Long short-term memory.
\newblock \emph{Neural Computation}, 9\penalty0 (8), 1997.

\bibitem[Horgan et~al.(2018)Horgan, Quan, Budden, Barth-Maron, Hessel,
  Van~Hasselt, and Silver]{horgan2018distributed}
Dan Horgan, John Quan, David Budden, Gabriel Barth-Maron, Matteo Hessel, Hado
  Van~Hasselt, and David Silver.
\newblock Distributed prioritized experience replay.
\newblock \emph{Proceedings of the International Conference on Learning
  Representations (ICLR)}, 2018.

\bibitem[Hu et~al.(2018)Hu, Shen, and Sun]{hu2018senet}
Jie Hu, Li~Shen, and Gang Sun.
\newblock Squeeze-and-excitation networks.
\newblock In \emph{Proceedings of IEEE Conference on Computer Vision and
  Pattern Recognition (CVPR)}, 2018.

\bibitem[James et~al.(2020)James, Ma, Rovick~Arrojo, and
  Davison]{james2019rlbench}
Stephen James, Zicong Ma, David Rovick~Arrojo, and Andrew~J. Davison.
\newblock Rlbench: The robot learning benchmark \& learning environment.
\newblock \emph{IEEE Robotics and Automation Letters}, 2020.

\bibitem[Jia et~al.(2018)Jia, Song, He, Wang, Rong, Zhou, Xie, Guo, Yang, Yu,
  et~al.]{jia2018highly}
Xianyan Jia, Shutao Song, Wei He, Yangzihao Wang, Haidong Rong, Feihu Zhou,
  Liqiang Xie, Zhenyu Guo, Yuanzhou Yang, Liwei Yu, et~al.
\newblock Highly scalable deep learning training system with mixed-precision:
  Training {ImageNet} in four minutes.
\newblock \emph{arXiv:1807.11205}, 2018.

\bibitem[Kempka et~al.(2016)Kempka, Wydmuch, Runc, Toczek, and
  Ja{\'s}kowski]{kempka2016vizdoom}
Micha{\l} Kempka, Marek Wydmuch, Grzegorz Runc, Jakub Toczek, and Wojciech
  Ja{\'s}kowski.
\newblock Vizdoom: A doom-based ai research platform for visual reinforcement
  learning.
\newblock In \emph{IEEE Conference on Computational Intelligence and Games},
  2016.

\bibitem[Keskar et~al.(2017)Keskar, Mudigere, Nocedal, Smelyanskiy, and
  Tang]{keskar2016large}
Nitish~Shirish Keskar, Dheevatsa Mudigere, Jorge Nocedal, Mikhail Smelyanskiy,
  and Ping Tak~Peter Tang.
\newblock On large-batch training for deep learning: Generalization gap and
  sharp minima.
\newblock \emph{Proceedings of the International Conference on Learning
  Representations (ICLR)}, 2017.

\bibitem[{Khronos Group}(2017)]{khronos2017vulkan}
{Khronos Group}.
\newblock The {Vulkan} specification.
\newblock 2017.

\bibitem[Kingma \& Ba(2015)Kingma and Ba]{kingma2016adam}
Diederik~P Kingma and Jimmy Ba.
\newblock Adam: A method for stochastic optimization.
\newblock \emph{Proceedings of the International Conference on Learning
  Representations (ICLR)}, 2015.

\bibitem[Kolve et~al.(2017)Kolve, Mottaghi, Han, VanderBilt, Weihs, Herrasti,
  Gordon, Zhu, Gupta, and Farhadi]{ai2thor}
Eric Kolve, Roozbeh Mottaghi, Winson Han, Eli VanderBilt, Luca Weihs, Alvaro
  Herrasti, Daniel Gordon, Yuke Zhu, Abhinav Gupta, and Ali Farhadi.
\newblock {AI2-THOR}: An interactive {3D} environment for visual {AI}.
\newblock \emph{arXiv:1712.05474}, 2017.

\bibitem[Küttler et~al.(2019)Küttler, Nardelli, Lavril, Selvatici, Sivakumar,
  Rocktäschel, and Grefenstette]{kttler2019torchbeast}
Heinrich Küttler, Nantas Nardelli, Thibaut Lavril, Marco Selvatici, Viswanath
  Sivakumar, Tim Rocktäschel, and Edward Grefenstette.
\newblock Torchbeast: A pytorch platform for distributed rl.
\newblock \emph{arXiv:1910.03552}, 2019.

\bibitem[Lee et~al.(2019)Lee, Hu, Yang, Yin, and Lim]{lee2019ikea}
Youngwoon Lee, Edward~S Hu, Zhengyu Yang, Alex Yin, and Joseph~J Lim.
\newblock {IKEA} furniture assembly environment for long-horizon complex
  manipulation tasks.
\newblock \emph{arXiv:1911.07246}, 2019.

\bibitem[Liang et~al.(2018)Liang, Liaw, Nishihara, Moritz, Fox, Goldberg,
  Gonzalez, Jordan, and Stoica]{liang2018rllib}
Eric Liang, Richard Liaw, Robert Nishihara, Philipp Moritz, Roy Fox, Ken
  Goldberg, Joseph~E. Gonzalez, Michael~I. Jordan, and Ion Stoica.
\newblock {RLlib}: Abstractions for distributed reinforcement learning.
\newblock In \emph{International Conference on Machine Learning ({ICML})},
  2018.

\bibitem[Loshchilov \& Hutter(2018)Loshchilov and
  Hutter]{loshchilov2018decoupled}
Ilya Loshchilov and Frank Hutter.
\newblock Decoupled weight decay regularization.
\newblock \emph{Proceedings of the International Conference on Learning
  Representations (ICLR)}, 2018.

\bibitem[OpenAI et~al.(2019)OpenAI, Berner, Brockman, Chan, Cheung, Debiak,
  Dennison, Farhi, Fischer, Hashme, Hesse, Józefowicz, Gray, Olsson, Pachocki,
  Petrov, de~Oliveira~Pinto, Raiman, Salimans, Schlatter, Schneider, Sidor,
  Sutskever, Tang, Wolski, and Zhang]{openai2019dota}
OpenAI, Christopher Berner, Greg Brockman, Brooke Chan, Vicki Cheung,
  Przemyslaw Debiak, Christy Dennison, David Farhi, Quirin Fischer, Shariq
  Hashme, Chris Hesse, Rafal Józefowicz, Scott Gray, Catherine Olsson, Jakub
  Pachocki, Michael Petrov, Henrique~Pondé de~Oliveira~Pinto, Jonathan Raiman,
  Tim Salimans, Jeremy Schlatter, Jonas Schneider, Szymon Sidor, Ilya
  Sutskever, Jie Tang, Filip Wolski, and Susan Zhang.
\newblock Dota 2 with large scale deep reinforcement learning.
\newblock 2019.
\newblock URL \url{https://arxiv.org/abs/1912.06680}.

\bibitem[Petrenko et~al.(2020)Petrenko, Huang, Kumar, Sukhatme, and
  Koltun]{petrenko2020sample}
Aleksei Petrenko, Zhehui Huang, Tushar Kumar, Gaurav Sukhatme, and Vladlen
  Koltun.
\newblock Sample factory: Egocentric {3D} control from pixels at 100000 fps
  with asynchronous reinforcement learning.
\newblock \emph{Proceedings of the International Conference on Machine Learning
  (ICML)}, 2020.

\bibitem[Ridnik et~al.(2020)Ridnik, Lawen, Noy, and
  Friedman]{ridnik2020tresnet}
Tal Ridnik, Hussam Lawen, Asaf Noy, and Itamar Friedman.
\newblock Tresnet: High performance gpu-dedicated architecture.
\newblock \emph{arXiv:2003.13630}, 2020.

\bibitem[Savva et~al.(2017)Savva, Chang, Dosovitskiy, Funkhouser, and
  Koltun]{savva2017minos}
Manolis Savva, Angel~X. Chang, Alexey Dosovitskiy, Thomas Funkhouser, and
  Vladlen Koltun.
\newblock {MINOS}: Multimodal indoor simulator for navigation in complex
  environments.
\newblock \emph{arXiv:1712.03931}, 2017.

\bibitem[Savva et~al.(2019)Savva, Kadian, Maksymets, Zhao, Wijmans, Jain,
  Straub, Liu, Koltun, Malik, Parikh, and Batra]{habitat19iccv}
Manolis Savva, Abhishek Kadian, Oleksandr Maksymets, Yili Zhao, Erik Wijmans,
  Bhavana Jain, Julian Straub, Jia Liu, Vladlen Koltun, Jitendra Malik, Devi
  Parikh, and Dhruv Batra.
\newblock Habitat: {A} {P}latform for {E}mbodied {AI} {R}esearch.
\newblock In \emph{Proceedings of IEEE International Conference on Computer
  Vision (ICCV)}, 2019.

\bibitem[Schulman et~al.(2016)Schulman, Moritz, Levine, Jordan, and
  Abbeel]{schulman2016high}
John Schulman, Philipp Moritz, Sergey Levine, Michael Jordan, and Pieter
  Abbeel.
\newblock High-dimensional continuous control using generalized advantage
  estimation.
\newblock \emph{Proceedings of the International Conference on Learning
  Representations (ICLR)}, 2016.

\bibitem[Schulman et~al.(2017)Schulman, Wolski, Dhariwal, Radford, and
  Klimov]{schulman2017ppo}
John Schulman, Filip Wolski, Prafulla Dhariwal, Alec Radford, and Oleg Klimov.
\newblock Proximal policy optimization algorithms.
\newblock \emph{arXiv:1707.06347}, 2017.

\bibitem[Silver et~al.(2017)Silver, Schrittwieser, Simonyan, Antonoglou, Huang,
  Guez, Hubert, Baker, Lai, Bolton, et~al.]{silver2017mastering}
David Silver, Julian Schrittwieser, Karen Simonyan, Ioannis Antonoglou, Aja
  Huang, Arthur Guez, Thomas Hubert, Lucas Baker, Matthew Lai, Adrian Bolton,
  et~al.
\newblock Mastering the game of {Go} without human knowledge.
\newblock \emph{Nature}, 550\penalty0 (7676), 2017.

\bibitem[Snook(2000)]{navmesh}
Greg Snook.
\newblock Simplified 3d movement and pathfinding using navigation meshes.
\newblock In Mark DeLoura (ed.), \emph{Game Programming Gems}, pp.\  288--304.
  Charles River Media, 2000.

\bibitem[Stooke \& Abbeel(2019)Stooke and Abbeel]{stooke2019rlpyt}
Adam Stooke and Pieter Abbeel.
\newblock rlpyt: A research code base for deep reinforcement learning in
  pytorch.
\newblock \emph{arXiv:1909.01500}, 2019.

\bibitem[Vinyals et~al.(2019)Vinyals, Babuschkin, Czarnecki, Mathieu, Dudzik,
  Chung, Choi, Powell, Ewalds, Georgiev, et~al.]{alphastarblog}
Oriol Vinyals, Igor Babuschkin, Wojciech~M. Czarnecki, Micha{\"e}l Mathieu,
  Andrew Dudzik, Junyoung Chung, David~H. Choi, Richard Powell, Timo Ewalds,
  Petko Georgiev, et~al.
\newblock Grandmaster level in {StarCraft II} using multi-agent reinforcement
  learning.
\newblock \emph{Nature}, 575\penalty0 (7782), 2019.

\bibitem[Weihs et~al.(2020)Weihs, Salvador, Kotar, Jain, Zeng, Mottaghi, and
  Kembhavi]{AllenAct}
Luca Weihs, Jordi Salvador, Klemen Kotar, Unnat Jain, Kuo-Hao Zeng, Roozbeh
  Mottaghi, and Aniruddha Kembhavi.
\newblock Allenact: A framework for embodied ai research.
\newblock \emph{arXiv}, 2020.

\bibitem[Wijmans et~al.(2020)Wijmans, Kadian, Morcos, Lee, Essa, Parikh, Savva,
  and Batra]{ddppo}
Erik Wijmans, Abhishek Kadian, Ari Morcos, Stefan Lee, Irfan Essa, Devi Parikh,
  Manolis Savva, and Dhruv Batra.
\newblock {DD-PPO}: Learning near-perfect pointgoal navigators from 2.5 billion
  frames.
\newblock In \emph{Proceedings of the International Conference on Learning
  Representations (ICLR)}, 2020.

\bibitem[Xia et~al.(2018)Xia, Zamir, He, Sax, Malik, and
  Savarese]{xia2018gibson}
Fei Xia, Amir~R Zamir, Zhiyang He, Alexander Sax, Jitendra Malik, and Silvio
  Savarese.
\newblock Gibson env: Real-world perception for embodied agents.
\newblock In \emph{Proceedings of IEEE Conference on Computer Vision and
  Pattern Recognition (CVPR)}, 2018.
\newblock {Gibson} dataset license agreement available at
  \url{https://storage.googleapis.com/gibson_material/Agreement%20GDS%2006-04-18.pdf}.

\bibitem[Xia et~al.(2020)Xia, Shen, Li, Kasimbeg, Tchapmi, Toshev,
  Mart{\'\i}n-Mart{\'\i}n, and Savarese]{xia2020interactive}
Fei Xia, William~B Shen, Chengshu Li, Priya Kasimbeg, Micael~Edmond Tchapmi,
  Alexander Toshev, Roberto Mart{\'\i}n-Mart{\'\i}n, and Silvio Savarese.
\newblock Interactive {Gibson} benchmark: A benchmark for interactive
  navigation in cluttered environments.
\newblock \emph{IEEE Robotics and Automation Letters}, 5\penalty0 (2), 2020.

\bibitem[Xiang et~al.(2020)Xiang, Qin, Mo, Xia, Zhu, Liu, Liu, Jiang, Yuan,
  Wang, et~al.]{Xiang2020}
Fanbo Xiang, Yuzhe Qin, Kaichun Mo, Yikuan Xia, Hao Zhu, Fangchen Liu, Minghua
  Liu, Hanxiao Jiang, Yifu Yuan, He~Wang, et~al.
\newblock {SAPIEN}: A simulated part-based interactive environment.
\newblock In \emph{Proceedings of IEEE Conference on Computer Vision and
  Pattern Recognition (CVPR)}, 2020.

\bibitem[You et~al.(2017)You, Gitman, and Ginsburg]{you2017scaling}
Yang You, Igor Gitman, and Boris Ginsburg.
\newblock Scaling {SGD} batch size to {32K} for {ImageNet} training.
\newblock \emph{arXiv:1708.03888}, 2017.

\bibitem[You et~al.(2020)You, Li, Reddi, Hseu, Kumar, Bhojanapalli, Song,
  Demmel, Keutzer, and Hsieh]{you2020large}
Yang You, Jing Li, Sashank Reddi, Jonathan Hseu, Sanjiv Kumar, Srinadh
  Bhojanapalli, Xiaodan Song, James Demmel, Kurt Keutzer, and Cho-Jui Hsieh.
\newblock Large batch optimization for deep learning: Training {BERT} in 76
  minutes.
\newblock \emph{Proceedings of the International Conference on Learning
  Representations (ICLR)}, 2020.

\bibitem[Zeng et~al.(2020)Zeng, Florence, Tompson, Welker, Chien, Attarian,
  Armstrong, Krasin, Duong, Sindhwani, and Lee]{zeng2020transporter}
Andy Zeng, Pete Florence, Jonathan Tompson, Stefan Welker, Jonathan Chien,
  Maria Attarian, Travis Armstrong, Ivan Krasin, Dan Duong, Vikas Sindhwani,
  and Johnny Lee.
\newblock Transporter networks: Rearranging the visual world for robotic
  manipulation.
\newblock \emph{Conference on Robot Learning (CoRL)}, 2020.

\bibitem[Zhang et~al.(2019)Zhang, Dauphin, and Ma]{zhang2019fixup}
Hongyi Zhang, Yann~N Dauphin, and Tengyu Ma.
\newblock Fixup initialization: Residual learning without normalization.
\newblock \emph{Proceedings of the International Conference on Learning
  Representations (ICLR)}, 2019.

\end{thebibliography}
